\documentclass[times,twocolumn,final,authoryear]{elsarticle}

\usepackage{ycviu}
\usepackage{framed,multirow}

\usepackage{amssymb}
\usepackage{latexsym}

\usepackage{hyperref}
\usepackage{soul}
\usepackage{graphicx}
\usepackage{multirow}
\usepackage{float}
\usepackage{listings}
\usepackage{bbm}
\usepackage{pifont}

\newcommand{\myparagraph}[1]{\vspace{6pt}\noindent{\bf #1}}

\usepackage{url}
\usepackage{xcolor}
\definecolor{newcolor}{rgb}{.8,.349,.1}

\definecolor{lightgray}{gray}{0.4}
\definecolor{verylightgray}{gray}{0.7}
\definecolor{veryverylightgray}{gray}{0.9}

\definecolor{darkgreen}{rgb}{0, 0.4, 0}
\definecolor{darkred}{rgb}{0.7, 0, 0}

\journal{Computer Vision and Image Understanding}

\begin{document}

\ifpreprint
  \setcounter{page}{1}
\else
  \setcounter{page}{1}
\fi

\begin{frontmatter}

\title{Generating Novel Scene Compositions from Single Images and Videos}

\author[1]{Vadim \snm{Sushko}\corref{cor1}} 
\cortext[cor1]{Corresponding author: Vadim Sushko
  }
\author[2]{Dan \snm{Zhang}$^{\textrm{\footnotesize{a}},}$}
\author[4]{Juergen \snm{Gall}$^{\textrm{\footnotesize{c}},}$}
\author[2]{Anna \snm{Khoreva}$^{\textrm{\footnotesize{a}},}$}

\address[1]{Bosch Center for Artificial Intelligence, Renningen, 71272, Germany}
\address[2]{University of T{\"u}bingen, T{\"u}bingen, 72070, Germany}
\address[3]{Lamarr Institute for Machine Learning and Artificial Intelligence, Germany}
\address[4]{University of Bonn, Bonn, 53115, Germany}

\received{July 2023}
\finalform{...}
\accepted{...}
\availableonline{...}
\communicated{...}

\begin{abstract}
Given a large dataset for training, generative adversarial networks (GANs) can achieve remarkable performance for the image synthesis task. However, training GANs in extremely low data regimes remains a challenge as overfitting often occurs, leading to memorization or training divergence.
In this work, we introduce SIV-GAN, an unconditional generative model that can generate new scene compositions from a single training image or a single video clip.   
We propose a two-branch discriminator architecture with content and layout branches that are designed to judge internal content and scene layout realism separately from each other. This discriminator design enables the synthesis of visually plausible, novel compositions of a scene with varying content and layout while preserving the context of the original sample. Compared to previous single-image GANs, our model generates more diverse images of higher quality while not being restricted to a single image setting. We further introduce a new challenging task of learning from a few frames of a single video. In this training setup the training images are highly similar to each other, which makes it difficult for prior GAN models to achieve a synthesis of both high quality and diversity. 
\end{abstract}


\end{frontmatter}



\section{Introduction}\label{sec:introduction}

In recent years, image generation models, such as generative adversarial networks (GANs) \citep{Brock2019,Karras2019stylegan2,karras2021alias,sauer2023stylegan} and diffusion models (DMs) \citep{ho2020denoising,nichol2021improved,dhariwal2021diffusion,rombach2022high}, have achieved remarkable advancements. These models have demonstrated impressive capabilities for image synthesis when trained on large and diverse datasets comprising thousands of images. For many real-world applications, however, the collection of extensive datasets is often unfeasible due to various constraints. These constraints may arise from privacy concerns, the rarity of objects or events, dangerous environments, or the nature of the application design itself. For instance, there may exist only a single image depicting a specific manufacturing defect or just one video capturing a traffic accident recorded under extreme conditions. Enabling the learning of generative models in such scenarios has great potential for practical applications. 

\begin{figure*}[h]  
	
\begin{centering}
\setlength{\tabcolsep}{0.1em}
\renewcommand{\arraystretch}{0}
\par\end{centering}
\begin{centering}
\vspace{-0.2em}
\hfill{}%
\begin{tabular}{@{\hskip 0.05in}c@{\hskip 0.08in}c@{\hskip -0.05in}}

Training image and samples generated from \emph{a single image} \tabularnewline

\includegraphics[width=0.16\linewidth, height=0.082\textheight]{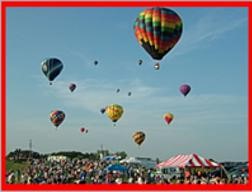}
\includegraphics[width=0.16\linewidth, height=0.082\textheight]{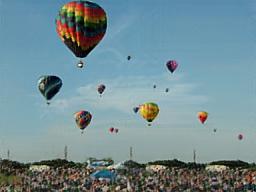}
\includegraphics[width=0.16\linewidth, height=0.082\textheight]{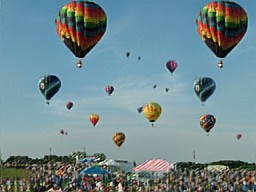}
\includegraphics[width=0.16\linewidth, height=0.082\textheight]{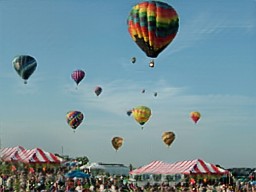}
\includegraphics[width=0.16\linewidth, height=0.082\textheight]{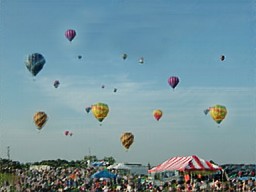}
\includegraphics[width=0.16\linewidth, height=0.082\textheight]{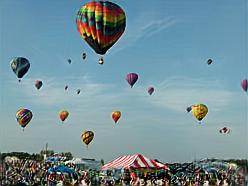}
\tabularnewline

\includegraphics[width=0.16\linewidth, height=0.082\textheight]{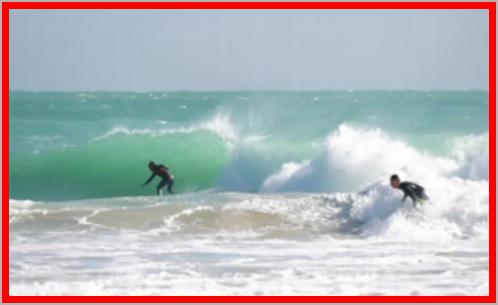}
\includegraphics[width=0.16\linewidth, height=0.082\textheight]{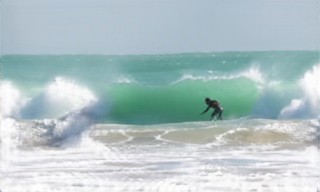}
\includegraphics[width=0.16\linewidth, height=0.082\textheight]{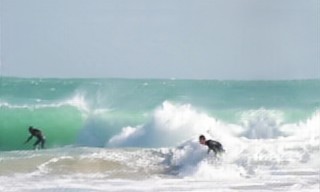}
\includegraphics[width=0.16\linewidth, height=0.082\textheight]{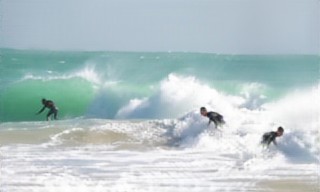}
\includegraphics[width=0.16\linewidth, height=0.082\textheight]{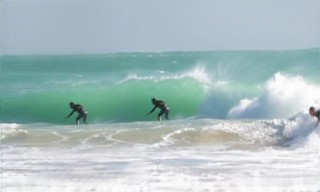}
\includegraphics[width=0.16\linewidth, height=0.082\textheight]{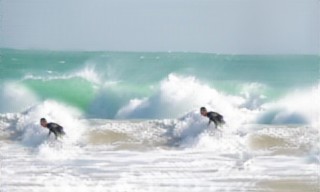}
\tabularnewline  \vspace{-2.5ex}
\tabularnewline

Training video and samples generated from \emph{a single video} \tabularnewline
\includegraphics[width=0.4845\linewidth, height=0.082\textheight]{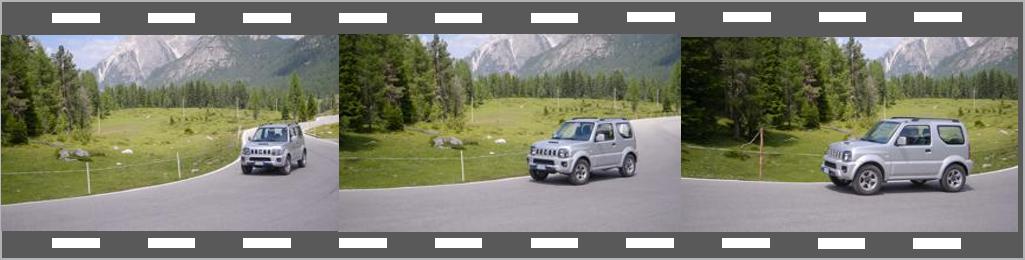} ~
\includegraphics[width=0.158\linewidth, height=0.082\textheight]{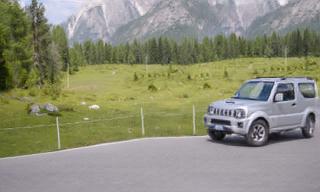} 
\includegraphics[width=0.158\linewidth, height=0.082\textheight]{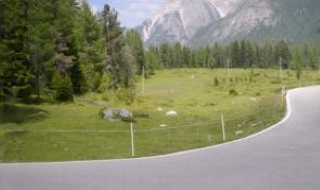}
\includegraphics[width=0.158\linewidth, height=0.082\textheight]{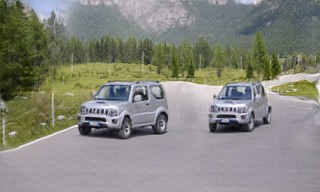}
 \tabularnewline

\includegraphics[width=0.4845\linewidth, height=0.082\textheight]{figures/teaser/park/ref}  ~
\includegraphics[width=0.158\linewidth, height=0.082\textheight]{figures/teaser/park/1}
\includegraphics[width=0.158\linewidth, height=0.082\textheight]{figures/teaser/park/32}
\includegraphics[width=0.158\linewidth, height=0.082\textheight]{figures/teaser/park/25}

\end{tabular}\hfill{}
\par\end{centering}
\setcounter{figure}{0}
\vspace{-1.1em}
\caption{\label{fig:teaser} Images generated by SIV-GAN. Our model successfully operates in extremely low data regimes, generating new scene compositions with varying content and layout from a single image (first two rows) or a single video (last two rows). For example, given a single surfing image, it can synthesize layouts with a different position and configuration of waves and change the number of surfers; and from a single video with a car on the road, SIV-GAN generates images without a car or with two cars. Original training samples are shown in red or grey frames.}
\end{figure*} 


Recent studies in image synthesis from extremely limited data have predominantly employed transfer learning, where a model is first pre-trained on a large dataset and subsequently fine-tuned using very small datasets \citep{ojha2021few,xiao2022few,giannone2022few}. A notable drawback of transfer learning in image synthesis is its requirement for pre-training datasets that are similar to the domain of interest, as fine-tuning a generative model pre-trained on a dissimilar dataset often leads to poor results, as shown in \citep{Zhao2020DifferentiableAF,ojha2021few}. 
This limitation poses a significant challenge for numerous applications where large datasets are lacking. To address this bottleneck and expand the applicability of generative models, we propose a novel GAN model that can be trained effectively from scratch using very limited data, eliminating the need for pre-training.

Previous research on GAN training from scratch has primarily focused on two scenarios with limited data: learning from a single image \citep{Shaham2019SinGANLA,hinz2021improved} (Fig.~\ref{fig:teaser}, rows 1-2), or from few-shot datasets comprising around 100 diverse images or more \citep{anonymous2021towards}. However, our experiments reveal that the latter methods are still constrained by the number and diversity of available training images. Notably, when trained on few-shot datasets consisting of very similar images, the few-shot model FastGAN \citep{anonymous2021towards} exhibits a significant drop in performance. To further explore the limitations of existing models, we propose a new task: generating images from a single video. In this task, the training data consists of 60-100 frames extracted from a short video clip lasting 2-10 seconds (see Fig.~\ref{fig:teaser}, rows 3-4). Similarly to \cite{anonymous2021towards}, our objective is to generate diverse images, but not temporally-coherent videos. Compared to a few-shot dataset containing 60-100 images, frames extracted from the same video exhibit a lower diversity between images due to the high correlation between adjacent video frames. Our experiments demonstrate that both single-image and few-shot GAN models struggle with this data regime, thereby establishing it as an interesting evaluation benchmark that can expand the applications of generative models to novel image domains.

Unlike previous single-image and few-shot GANs, our proposed model, called SIV-GAN (\textbf{S}ingle \textbf{I}mage and \textbf{V}ideo GAN), addresses the difficulties associated with training from scratch in the above extremely limited data regimes. By utilizing just one or several very similar images, our model generates diverse images with new scene compositions, e.g., rearranging objects within scenes or modifying their shape and size, simply by resampling the input noise. We compare our model to single-image~\citep{Shaham2019SinGANLA,Hinz2020ImprovedTF} and few-shot GANs~\citep{anonymous2021towards}, as well as to the diffusion-based fine-tuning approach DreamBooth \citep{ruiz2023dreambooth}. 


A preliminary version of this work has been published in \citep{sushko2021one}. Compared to \citep{sushko2021one}, we provide a more detailed discussion and description of the approach and a thorough experimental evaluation and ablations studies. The ablation studies analyse the impact of diversity regularization, design choices, and the impact of the average similarity of training images.    

\section{Related work}
\label{sec:related_work}

The goal of this work is to train a generative model to synthesize diverse and high-quality images in extremely low data regimes from scratch. Among the two current dominant paradigms for image generation, GANs and diffusion models, only GAN approaches have been employed to tackle unconditional training from one or just a few images without using pre-training. On the contrary, most recent diffusion models still require large datasets, e.g., more than 5000 images for successful training \citep{wang2023patch}, or need pre-training \citep{giannone2022few,moon2022fine, ruiz2023dreambooth}. Therefore, our focus in this study is on GANs. The related literature to our work encompasses two types of models: single-image and few-shot GAN approaches.

\begin{figure*}[t]
\begin{centering}
\setlength{\tabcolsep}{0.1em}
\renewcommand{\arraystretch}{0}
\par\end{centering}
\begin{centering}
\hfill{}%
\begin{tabular}{c}

\includegraphics[width=0.99\linewidth, height=0.25\linewidth]{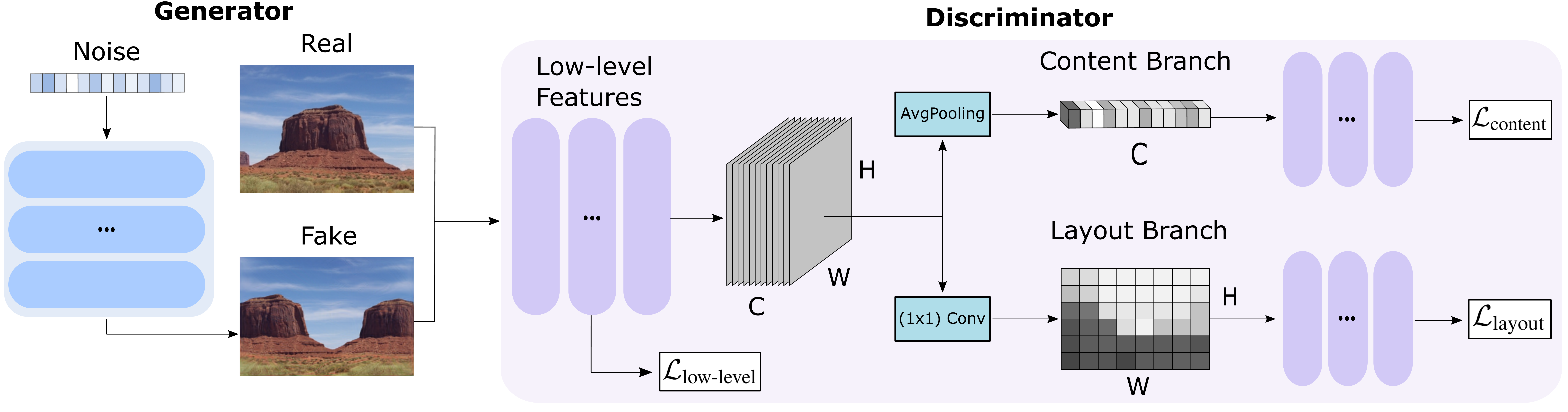} \tabularnewline

\end{tabular}\hfill{}
\par\end{centering}
\vspace{-1.4em}
\caption{\label{fig:model_over} SIV-GAN architecture. Two separate discriminator branches judge the image content separately from the scene layout realism and thus enable the generator to produce images with varying content and global layouts. Before the branching, a low-level feature extractor is trained via a separate low-level loss, enabling to learn the low-level image realism and to build relevant representations for the content and layout branches.	
%
}
\vspace{-1.4em}
\end{figure*}

\myparagraph{Single-image GANs.} 
The challenge of training GANs from limited data is the problem of overfitting, which leads to memorization issues and training divergence~\citep{Karras2020TrainingGA}. As one approach to mitigate the memorization issues, single-image GAN models proposed not to learn the statistics of the whole given image, but only the distribution of its patches. For example, this can be achieved by employing a cascade of multi-scale patch-GANs~\citep{isola2017image} trained in multiple stages. For example, SinGAN \citep{Shaham2019SinGANLA} employed an unconditional GAN to produce images of arbitrary size from noise, and used a multi-stage training scheme to learn the multi-scale patch distribution of an image. ConSinGAN \citep{Hinz2020ImprovedTF} improved SinGAN by rescaling the multi-stage training and training several stages concurrently, which enabled reducing the model size and made the training more efficient.

Contrary to the above GANs, our model is trained in a single stage, and is designed to learn not only the internal patch-based distribution of an image, but also to capture high-level content, such as scene layouts or appearance of objects. As shown by our experiments, this enables generating more diverse, higher quality synthesis from a single image, as well as to learn from multiple similar images.

\myparagraph{Few-shot GANs.} 
Another line of work focused on improving the stability of GANs when trained with few images. To prevent the GAN discriminator from overfitting, \cite{Karras2020TrainingGA}, \cite{Zhao2020ImageAF}, and \cite{Zhao2020DifferentiableAF} explored differentiable data augmentations (DA) for both real and generated images. Since then, \cite{tseng2021regularizing}, \cite{yang2021data}, \cite{jiang2021deceive}, and \cite{chen2021data} proposed various alternative approaches, showing them to be complementary to DA techniques. These works used limited, but still relatively large training sets ($\geqslant$ $1$k images) compared to the few-shot setting and our proposed single video regimes ($\leqslant$ $100$). As our experiments demonstrate (see Sec.~\ref{sec:exp_ablations}), using DA alone is not enough to achieve good diversity when learning from a single image or video.

Another work, FastGAN \citep{anonymous2021towards}, proposed to train an unconditional GAN from scratch on around 100 images. This model avoids overfitting thanks to the proposed skip-layer channel-wise excitation module in the generator and a self-supervised discriminator. However, FastGAN still struggles to successfully learn from a single image and even from a single video: although our proposed single video setting has a similar number of training frames to a standard few-shot dataset ($\sim$100 samples), it provides much less variability in training data due to a high correlation of adjacent frames, which leads to memorization issues, as will be discussed in Sec.~\ref{sec:exp_sin_vid}.

\myparagraph{Content-style separation in GANs.} 
Our work is also related to prior research on the content-style separation in GANs. Prior works focused on disentangling content and style features for unconditional image synthesis \citep{wu2019disentangling}, image-to-image translation \citep{huang2018multimodal}, or image manipulation \citep{Park2020SwappingAF}. These models generate different images by separately varying the content and style vectors used for the generator. In this work, we differently explore the idea of separating the \textit{discriminator's} learning of image content and layout, and demonstrate its potential to mitigate memorization and induce diversity in extremely low data regimes.

\section{SIV-GAN}
\label{sec:method}


In this section, we present SIV-GAN, an unconditional GAN model that learns from a single image or a single video to generate new plausible compositions of a given scene with varying content and layout.  
The key ingredients of SIV-GAN are a novel design of a two-branch discriminator (Sec.~\ref{sec:method_discriminator}) and a diversity regularization introduced for synthesis in single data instance regimes (Sec.~\ref{sec:method_DR}).


\subsection{Content-layout discriminator} \label{sec:method_discriminator}

\begin{figure}[t]
\begin{centering}
\setlength{\tabcolsep}{0.0em}
\renewcommand{\arraystretch}{0}
\par\end{centering}
\begin{centering}
		\begin{tabular}{@{\hskip -0.05in}c@{\hskip -0.02in}c@{\hskip 0.03in}c@{}}

			&\multicolumn{2}{c}{\hspace{-0.1in} \small ~SinGAN, lowest resolution of 20px}  
			\tabularnewline	

			\multirow{-3}{*}{\begin{tabular}{c}  Training image \\ 	\includegraphics[width=0.31\linewidth, height=0.065\textheight]{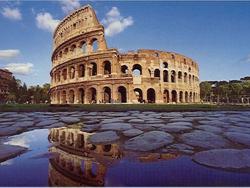}\end{tabular}} &

			\includegraphics[width=0.31\linewidth, height=0.065\textheight]{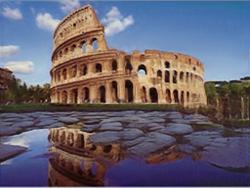} & 
			\includegraphics[width=0.32\linewidth, height=0.065\textheight]{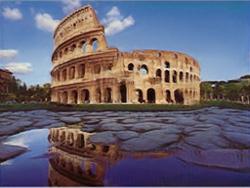} 	\tabularnewline

			& \multicolumn{2}{c}{ \hspace{-0.1in}  \small ~SinGAN, lowest resolution of 35px} \tabularnewline
			
			& \includegraphics[width=0.32\linewidth, height=0.065\textheight]{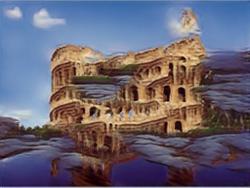} & 
			\includegraphics[width=0.32\linewidth, height=0.065\textheight]{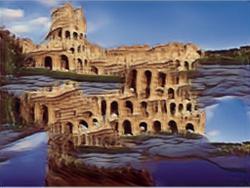}
			
			\tabularnewline

		\end{tabular}\hfill{}
		\par\end{centering}
	\vspace{-0.6em}
	\caption{\label{fig:singan_failures} Limitation of the multi-stage training of single image patch-GAN methods. 
		As the finer generation stages cannot correct the layout decision made at the coarser scales, the model produces images of very low diversity (first row) or lacking global coherency (last row) depending on the lowest resolution size.  
	}
	\vspace{-1.0em}
\end{figure}

One challenge of training GANs in single data instance regimes is the problem of overfitting to original samples. In many cases the model can simply memorize the original training images and their augmented versions used during training. To avoid this memorization effect, \cite{Shaham2019SinGANLA} and \cite{Hinz2020ImprovedTF} proposed to learn an internal patch-based distribution of a single image by using a hierarchy of patch-GANs \citep{isola2017image} at different image scales. As the employed patch-GANs have small receptive fields and limited capacity, they are prevented from memorizing the full image. 
However, the downside of training each scale of the patch-GANs in a separate stage is that any layout decisions made by the coarser scale generators cannot be corrected at later,
finer generation stages. Thus, the quality and diversity of generated images are highly dependent on the chosen lowest resolution size.
This parameter needs
careful tuning for specific images at hand, otherwise image layouts
may lack diversity or lose global coherency (see Fig. \ref{fig:singan_failures}). 
Moreover, this approach does not generalize to learning from multiple images, as in the single video case (see Fig.~\ref{fig:qual_video}).

We therefore introduce an alternative solution to overcome the memorization effect but still to produce high-quality images.
We note that in order to produce realistic and diverse images, the generator should learn the appearance of objects and combine them in the image in a globally-coherent way. To this end, we propose a discriminator that judges the \textit{content} distribution of a given image separately from its \textit{layout} realism. 
To achieve the disentanglement, we design a two-branch discriminator architecture with separate content and layout branches.
Note that the branching of the discriminator happens after intermediate layers; this is done in order to learn relevant representations for building the branches.
As seen from Fig.~\ref{fig:model_over}, our discriminator consists of the low-level feature extractor $D_{low-level}$, the content branch $D_{content}$, and the layout branch $D_{layout}$. For a given image $x$, the purpose of $D_{low-level}$ is to learn low-level features and to produce an image representation $F(x)= D_{low-level}(x)$ for the 
branches. Next, $D_{content}$ will judge the content of $F(x)$, irrespective from its spatial layout, while on the other hand $D_{layout}$ will inspect only the spatial information extracted from $F(x)$. 
Inspired by the attention modules of \cite{park2018bam} and \cite{woo2018cbam}, we implement the content-layout disentanglement by squeezing channels or spatial dimensions of the intermediate features $F(x)$. Note that afterwards the branches $D_{content}$ and $D_{layout}$ receive only limited information about the image from $F(x)$, preventing them from overfitting to the whole image, and thus mitigating the negative effect of memorizing the original image.




\myparagraph{Content branch.}
The content branch decision should be based upon the image content, i.e. the fidelity of objects composing the image, independent of their spatial location in the scene.
Let the feature map $F(x)$ have dimensions $H (\mathrm{height})\times W (\mathrm{width})\times C (\mathrm{channel})$. 
Note that the spatial dimensions $H\times W$ capture spatial information, while the channels $C$ encode the semantic representation. As we want the content branch to ignore the spatial location of objects, we apply global average pooling to aggregate the spatial information $H\times W$ across the channels $C$. The resulting feature map $F_{content}(x)$ has size $1\times1\times C$, which is then processed by several layers for further real/fake decision making. 
By removing the spatial information, $D_{content}(x)$ is induced to respond to content features encoded in different channels regardless of their spatial location (see Fig.~\ref{fig:qual_ablations_d}).

\myparagraph{Layout branch.}
The layout branch, in contrast, should assess the spatial location of objects in the scene, but not their specific appearance. Thus, the layout branch is designed to judge only the spatial information of $F(x)$, filtering out the content details. 
Since the layout information is encoded only in spatial dimensions $H \times W$, and not in channels $C$, we aggregate the channel information from $F(x)$ via a $(1\times 1)$ convolution with only one output channel, which forms a feature map $F_{layout}(x)$ with size $H \times W \times 1$.
This channel aggregation weakens the content representation but does not affect the spatial information. The $F_{layout}(x)$ features are further processed by several layers before a real/fake decision is made. 
As $D_{layout}(x)$ is designed to be sensitive only to the spatial representation of the input image, it learns to judge the realism of scene layouts (see Fig.~\ref{fig:qual_ablations_d}).


\myparagraph{Feature augmentation.}
The proposed two-branch discriminator prevents the memorization of training samples, enabling the generation of images with content and layouts different from the original sample.
To further improve the diversity of generated images, we propose to augment the content $F_{content}(x)$ and layout $F_{layout}(x)$ features of real images. For the single image setting this is done by mixing the features of two different augmentations of the original image, and for the single video setting by mixing the features of augmentations of two different video frames.
For two real samples $x_1$ and $x_2$, we apply a mixing transformation $F_{*}(x_1) = T_{mix}(F_{*}(x_1), F_{*}(x_2))$. We use two types of mixing: 1) For the layout branch, we sample a rectangular crop of $F_{layout}(x_2)$ and paste it on to $F_{layout}(x_1)$ at the same spatial location, similarly to CutMix \citep{Yun2019CutMixRS}. In contrast to CutMix, our approach augments features, not input images, and mixes only features of real images.
2) For the content branch, we sample a set of channels from $F_{content}(x_2)$ and copy their values to the corresponding channels of $F_{content}(x_1)$. As the channels encode semantic features of images, we expect the resulting augmented tensor to represent objects seen in two different images. We also found it useful to remove channels from $F_{content}(x)$, thus removing some object representations. 
For this, we sample a set of channels and drop out their values \citep{JMLR:v15:srivastava14a,tian2018dropfilter}. With the above augmentations, $D_{content}$ and $D_{layout}$ see significantly more variance in both the content and layout representations of real images, which prevents overfitting and improves the diversity of generated samples. The effect of feature augmentation (FA) is shown in Table \ref{table:main_ablation}.

\begin{figure*}[t]
\begin{centering}
\setlength{\tabcolsep}{0.0em}
\renewcommand{\arraystretch}{0}
\par\end{centering}
\begin{centering}
\begin{tabular}{@{\hskip -0.08in}c@{\hskip -0.02in}c@{\hskip 0.04in}c@{\hskip 0.09in}c@{\hskip 0.04in}c@{\hskip 0.09in}c@{\hskip 0.04in}c@{}}
& \multicolumn{2}{c}{SinGAN~\citep{Shaham2019SinGANLA}} &\multicolumn{2}{c}{ConSinGAN~\citep{Hinz2020ImprovedTF}}  & \multicolumn{2}{c}{SIV-GAN}  
\tabularnewline	
  \multirow{-3}{*}{\begin{tabular}{c} \hspace{-0.4em} Training image \\ 	\includegraphics[width=0.134\linewidth, height=0.055\textheight]{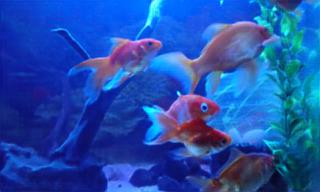} \end{tabular} }&
  \includegraphics[width=0.134\linewidth, height=0.055\textheight]{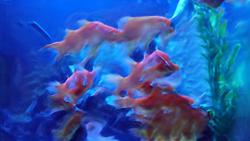} & 
  \includegraphics[width=0.134\linewidth, height=0.055\textheight]{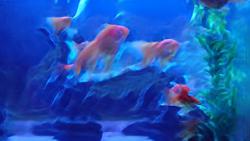} & 
  \includegraphics[width=0.134\linewidth, height=0.055\textheight]{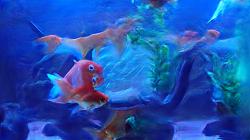} & 
  \includegraphics[width=0.134\linewidth, height=0.055\textheight]{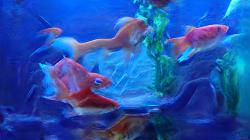} & 
  \includegraphics[width=0.134\linewidth, height=0.055\textheight]{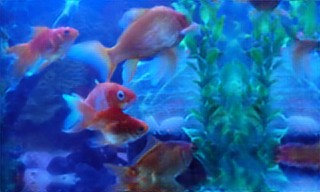} & 
  \includegraphics[width=0.134\linewidth, height=0.055\textheight]{figures/bank_sin_im/fish/41}
  \tabularnewline
  &
  \includegraphics[width=0.134\linewidth, height=0.055\textheight]{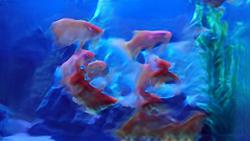} & 
  \includegraphics[width=0.134\linewidth, height=0.055\textheight]{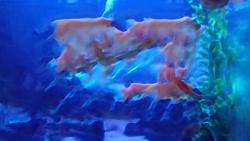} & 
  \includegraphics[width=0.134\linewidth, height=0.055\textheight]{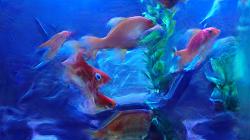} & 
  \includegraphics[width=0.134\linewidth, height=0.055\textheight]{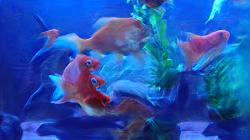} & 
  \includegraphics[width=0.134\linewidth, height=0.055\textheight]{figures/bank_sin_im/fish/42} & 
  \includegraphics[width=0.134\linewidth, height=0.055\textheight]{figures/bank_sin_im/fish/61} 
\vspace{0.3ex}
\tabularnewline


\multirow{-3}{*}{\begin{tabular}{c} \hspace{-0.4em} Training image \\\includegraphics[width=0.134\linewidth, height=0.055\textheight]{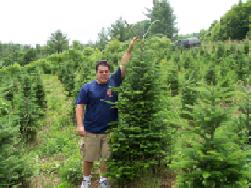} \end{tabular}} &
\includegraphics[width=0.134\linewidth, height=0.055\textheight]{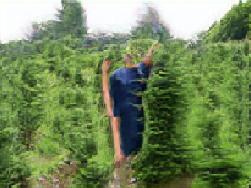} & 
\includegraphics[width=0.134\linewidth, height=0.055\textheight]{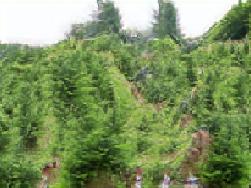} & 
\includegraphics[width=0.134\linewidth, height=0.055\textheight]{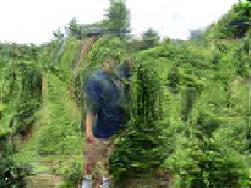} & 
\includegraphics[width=0.134\linewidth, height=0.055\textheight]{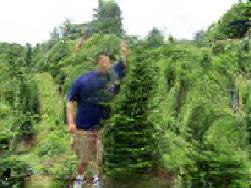} & 
\includegraphics[width=0.134\linewidth, height=0.055\textheight]{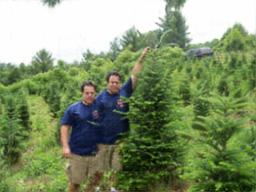} & 
\includegraphics[width=0.134\linewidth, height=0.055\textheight]{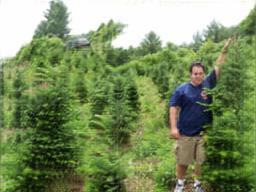}
\tabularnewline
&
\includegraphics[width=0.134\linewidth, height=0.055\textheight]{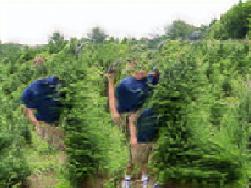} & 
\includegraphics[width=0.134\linewidth, height=0.055\textheight]{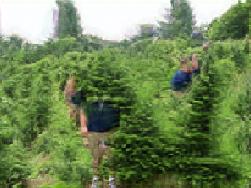} & 
\includegraphics[width=0.134\linewidth, height=0.055\textheight]{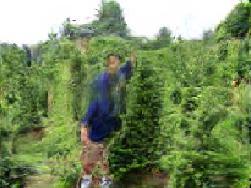} & 
\includegraphics[width=0.134\linewidth, height=0.055\textheight]{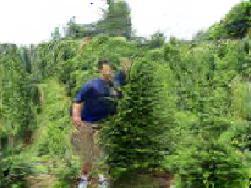} & 
\includegraphics[width=0.134\linewidth, height=0.055\textheight]{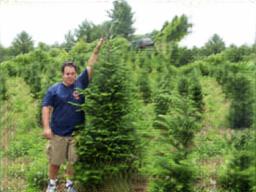} & 
\includegraphics[width=0.134\linewidth, height=0.055\textheight]{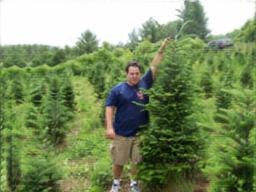} \vspace{0.9ex} \tabularnewline 

 & \multicolumn{2}{c}{SinGAN~\citep{Shaham2019SinGANLA}} &\multicolumn{2}{c}{FastGAN~\citep{anonymous2021towards}}  & \multicolumn{2}{c}{SIV-GAN} 
\tabularnewline	
\multirow{-3}{*}{\begin{tabular}{c} \hspace{-0.4em} Training image \\\includegraphics[width=0.134\linewidth, height=0.055\textheight]{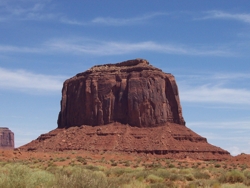} \end{tabular}} &
\includegraphics[width=0.134\linewidth, height=0.055\textheight]{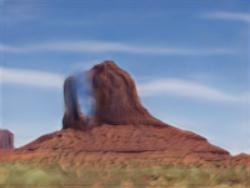} & 
\includegraphics[width=0.134\linewidth, height=0.055\textheight]{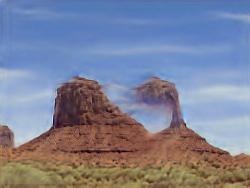} & 
\includegraphics[width=0.134\linewidth, height=0.055\textheight]{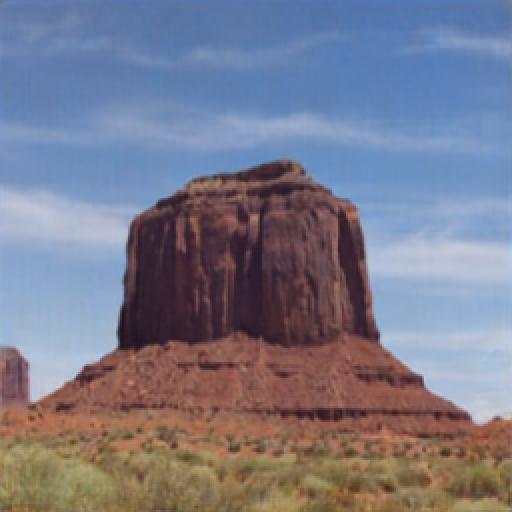} & 
\includegraphics[width=0.134\linewidth, height=0.055\textheight]{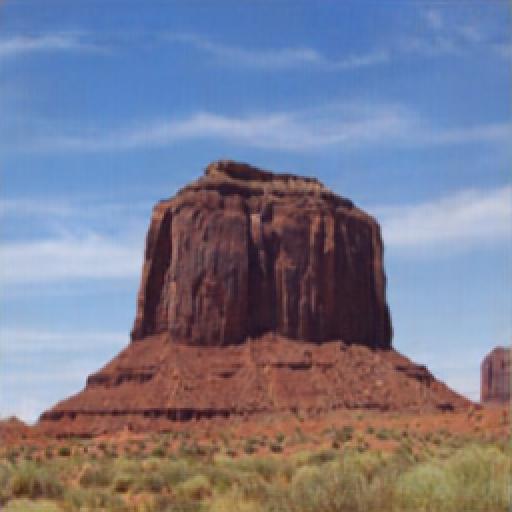} & 
\includegraphics[width=0.134\linewidth, height=0.055\textheight]{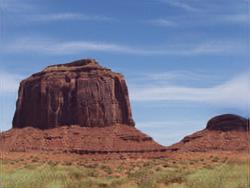} & 
\includegraphics[width=0.134\linewidth, height=0.055\textheight]{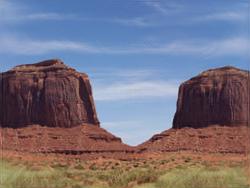} \tabularnewline 

& 
\includegraphics[width=0.134\linewidth, height=0.055\textheight]{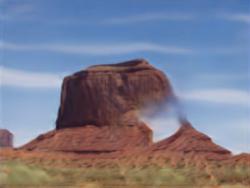} & 
\includegraphics[width=0.134\linewidth, height=0.055\textheight]{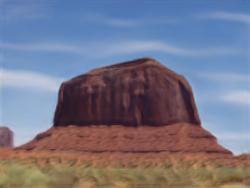} & 
\includegraphics[width=0.134\linewidth, height=0.055\textheight]{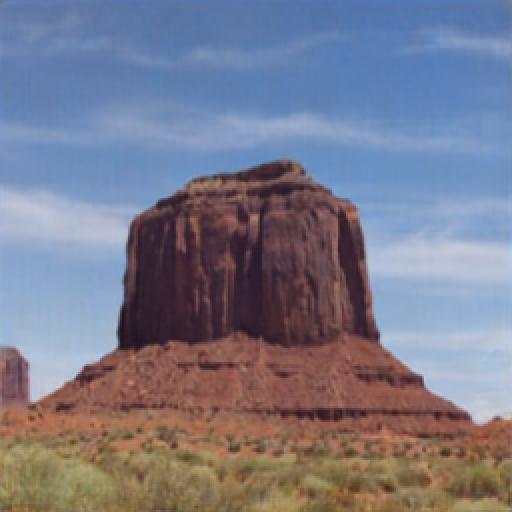} & 
\includegraphics[width=0.134\linewidth, height=0.055\textheight]{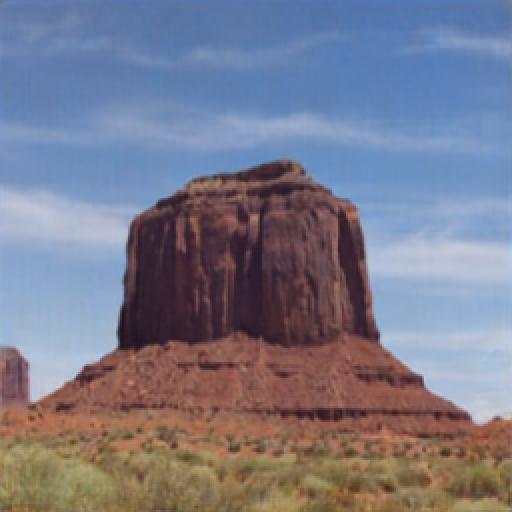} & 
\includegraphics[width=0.134\linewidth, height=0.055\textheight]{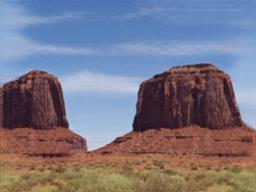} & 
\includegraphics[width=0.134\linewidth, height=0.055\textheight]{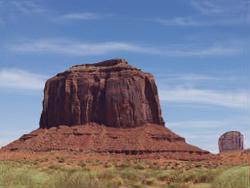}
\vspace{0.7ex}
\tabularnewline


\multirow{-3}{*}{\begin{tabular}{c} \hspace{-0.4em} Training image \\\includegraphics[width=0.134\linewidth, height=0.055\textheight]{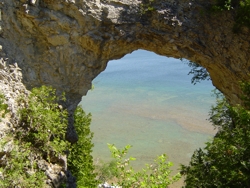} \end{tabular}} &
\includegraphics[width=0.134\linewidth, height=0.055\textheight]{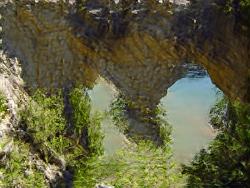} & 
\includegraphics[width=0.134\linewidth, height=0.055\textheight]{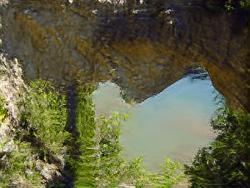} & 
\includegraphics[width=0.134\linewidth, height=0.055\textheight]{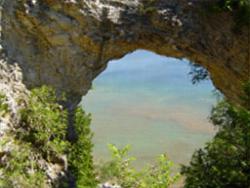} & 
\includegraphics[width=0.134\linewidth, height=0.055\textheight]{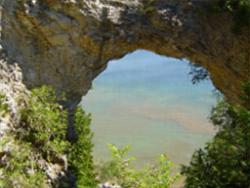} & 
\includegraphics[width=0.134\linewidth, height=0.055\textheight]{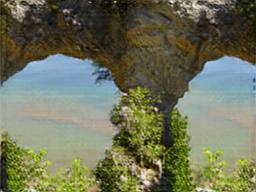} & 
\includegraphics[width=0.134\linewidth, height=0.055\textheight]{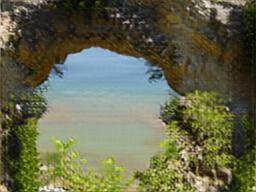}
\tabularnewline
&
\includegraphics[width=0.134\linewidth, height=0.055\textheight]{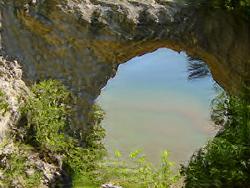} & 
\includegraphics[width=0.134\linewidth, height=0.055\textheight]{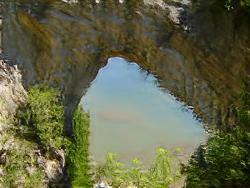} & 
\includegraphics[width=0.134\linewidth, height=0.055\textheight]{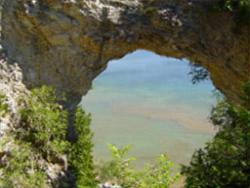} & 
\includegraphics[width=0.134\linewidth, height=0.055\textheight]{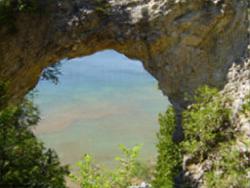} & 
\includegraphics[width=0.134\linewidth, height=0.055\textheight]{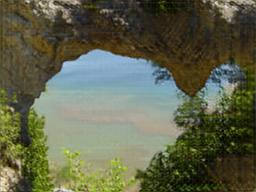} & 
\includegraphics[width=0.134\linewidth, height=0.055\textheight]{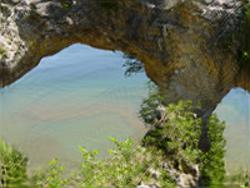}

\end{tabular}\hfill{}
\par\end{centering}
\vspace{-0.9em}
\caption{\label{fig:qual_image} 
	Visual comparison between models in the Single Image setting. Single image GANs are prone to shuffle patches incoherently (e.g. sky textures below horizon, perturbed fish contours), while the few-shot FastGAN suffers from memorization, reproducing only the original image or its flipped version. In contrast, SIV-GAN achieves both high quality and diversity, preserving the realism of image content and layout.
}
\vspace{-1.5em}
\end{figure*}

\myparagraph{Adversarial loss.}
To evaluate images at different scales, we design our discriminator to make a binary true/fake decision at each intermediate resolution. For each discriminator part $D_*$ ($D_{low-level}$, $D_{content}$, $D_{layout}$), the loss is computed by aggregating the contributions across all layers constituting the corresponding discriminator part:
\begin{equation}
\mathcal{L}_{D_{*}}  = \frac{1}{N_*}\sum\limits_{l=1}^{N_*} \mathcal{L}_{D_{*}^l}, \label{eq:loss_d_1}
\end{equation}
where $D_*^l$ is the $l$-th ResNet block of $D_*$, $N_{*}$ is the number of ResNet blocks used in $D_{*}$, and the loss $\mathcal{L}_{D^l_{*}}$ is the binary cross-entropy: 
\begin{equation}
\mathcal{L}_{D^l_{*}} = -\mathbb{E}_x[\log D^l_{*}(x)] - \mathbb{E}_z[\log (1-D^l_{*}(G(z)))].
\end{equation}
$D^l_{*}$ aims to distinguish between real $x$ and generated $G(z)$ images based on their corresponding features at block $l$, which captures either their low-level details, content, or layout at a certain resolution.
The overall adversarial loss for SIV-GAN is then computed by taking the decisions from the content branch $D_{content}$, the layout branch $D_{layout}$, and the low-level features of $D_{low-level}$:
\begin{equation}  \vspace{-5pt}
\mathcal{L}_{adv} (G,D)  = \mathcal{L}_{D_{content}} + \mathcal{L}_{D_{layout}} + 2 \mathcal{L}_{D_{low-level}}. \label{eq:loss_d_2}
\end{equation} 
As two $D$ branches operate on high-level image features, contrary to only one $D_{low-level}$ operating on low-level features, we weight $\mathcal{L}_{D_{low-level}}$ by factor two. This is done in order to properly balance the contributions of different feature scales and encourage the generation of images with good low-level details, plausible content, and coherent scene layouts. The effect of $\mathcal{L}_{D_{low-level}}$ is discussed in Sec. \ref{sec:exp_ablations}.



\subsection{Diversity regularization} \label{sec:method_DR}

\begin{table*}[t]

	\setlength{\tabcolsep}{0.30em}
	\renewcommand{\arraystretch}{1.00}
	\centering
 
	\caption{Comparison with other methods in the Single Image setting on Places and DAVIS-YFCC100M datasets.}
	\begin{tabular}{c|@{\hskip 0.02in}ccc|c|c|c||@{\hskip 0.02in}ccc|c|c|c}
		& \multicolumn{6}{c||@{\hskip 0.02in}}{\small{}Places}  & \multicolumn{6}{c}{\small{}DAVIS-YFCC100M}  
		\tabularnewline
		\multirow{2}{*}{\normalsize{} Method } &  \multicolumn{3}{c|}{\small{} SIFID~$\downarrow$ } & \multirow{2}{*}{\small{} LPIPS~$\uparrow$ } & {\small{} Pixel~$\uparrow$ } &  {\small{}Dist.~ } & \multicolumn{3}{c|}{\small{} SIFID~$\downarrow$ } & \multirow{2}{*}{\small{} LPIPS~$\uparrow$ } & {\small{} Pixel~$\uparrow$} &  {\small{}Dist.~ } \tabularnewline
		& \footnotesize{$\frac{H\times W}{4}$(best)} & \footnotesize{$\frac{H\times W}{4}$} & \footnotesize{$\frac{H\times W}{16}$} & & {\small{} Diversity } & to train & \footnotesize{$\frac{H\times W}{4}$(best)} & \footnotesize{$\frac{H\times W}{4}$} & \footnotesize{$\frac{H\times W}{16}$} & & {\small{} Diversity } & to train \tabularnewline
		
		\hline 	 \hline 	
		
		{\normalsize{} SinGAN} & \small{0.09}  & \small{0.15}  & \small{25.33} & \small{0.22} & \small{0.52} & \small{0.24} & \small{0.10} & \small{} 0.13 & \small{} 34.52 & \small{} 0.26 & \small{} 0.54 & \small{0.30} \tabularnewline
		
		{\normalsize{} ConSinGAN} & \small{0.06}  & \small{0.08}  & \small{23.45}   &  \small{0.24} & \small{0.50} & \small{0.25} & \small{0.08} & \small{} 0.09 & \small{} 27.33 & \small{} 0.29 & \small{} 0.59 & \small{0.31}  \tabularnewline
		
		{\normalsize{} FastGAN} & \small{0.11}  & \small{0.14} & \small{16.52} & \small{0.15} & \small{0.48} & \color{darkred} \small{0.08} & \small{0.10} & \small{} 0.13 & \small{} 19.48 & \small{} 0.18 & \small{} 0.49 & \color{darkred} \small{0.11}  \tabularnewline		
		
		{\normalsize{} SIV-GAN} & \textbf{\small{0.05}}  & \textbf{\small{0.06}} & \textbf{\small{12.12}} & \textbf{\small{0.28}} & \textbf{\small{0.57}}  & \small{0.31} & \textbf{\small{0.07}} & \small{} \textbf{0.08} & \small{} \textbf{16.30} & \small{} \textbf{0.33} & \small{} \textbf{0.66}  & \small{0.37}
		
	
		\end{tabular}
\label{table:comp_single_image} %

\end{table*}

To improve the variability among the generated images, we propose to add a diversity regularization (DR) loss term $\mathcal{L}_{DR}$ to the SIV-GAN objective.
Prior work \citep{Yang2019DiversitySensitiveCG,Zhao2020ImprovedCR, choi2020stargan} also proposed to use diversity regularization for GAN training, but mainly to avoid mode collapse and assuming the availability of a large training set. The regularization of \cite{Yang2019DiversitySensitiveCG} aimed to encourage the generator to produce different outputs depending on the input latent code in such a way that the generated samples with closer latent codes should look more similar to each other and vice versa. In contrast, our diversity regularization is tuned for synthesis from single data instance regimes. Assuming that in case of a single image or a single video we are operating in one semantic domain, the generator should produce images that are in-domain but more or less equally different from each other and substantially different from the original training sample. Thus, in such regimes the difference of generated images should not depend on the distance between their latent codes and we propose to encourage the generator
to produce perceptually different image samples independent of their distance in the latent space. Mathematically, the new diversity regularization is expressed as:
\begin{equation}
\mathcal{L}_{DR} (G) = \mathbb{E}_{z_1,z_2}\left[\frac{1}{L}\sum_{l=1}^{L}\|G^{l}(z_1)-G^{l}(z_2)\|\right], \label{loss_DR}
\end{equation}
where $\|\cdot\|$ denotes the $L1$ norm, $G^{l}(z)$ indicates a feature extracted after the $l$-th resolution block of the generator $G$ given input $z$, and $z_1, z_2$ are randomly sampled latent codes in the batch, i.e.\ $z_1,z_2{\sim}N(0,1)$.
By regularizing the generator to maximize Eq.~\ref{loss_DR}, we force it to produce diverse outputs for different latent codes $z$.
Note that, in contrast to \citep{Yang2019DiversitySensitiveCG}, \citep{Zhao2020ImprovedCR}, and \citep{choi2020stargan}, we compute the distance between samples in the feature space of the generator. Computing the distance in the feature space results in a more meaningful diversity among the generated images, as different layers of the generator capture different image semantics, inducing both high- and low-level diversity.
Computing the distance in the image space,
i.e. $\mathcal{L}_{DR} (G) = \|G(z_1)-G(z_2)\|$ as in \citep{choi2020stargan}, reduces the image diversity as shown in our experiments (see Table \ref{table:ablation_diversity}).
The overall SIV-GAN objective can be written as:
\begin{equation}
\max_{G} \min_{D} \hspace{0.5em} \mathcal{L}_{adv} (G,D) + \lambda \mathcal{L}_{DR} (G), \label{loss_full}
\end{equation}
where $\lambda$ controls the strength of the diversity regularization and $\mathcal{L}_{adv}$ is the adversarial loss in Eq.~\ref{eq:loss_d_2}.
The proposed diversity regularization is shown to be highly-effective for SIV-GAN,
while prior regularizations~\citep{Yang2019DiversitySensitiveCG,Zhao2020ImprovedCR, choi2020stargan} underperform in our experiments (see Table~\ref{table:ablation_diversity}).

\begin{table}[t]
\setlength{\tabcolsep}{0.4em}
\renewcommand{\arraystretch}{1.00}
\centering
\caption{Comparison in the Single Video setting on DAVIS-YFCC100M.}
\label{table:comp_single_video} %
	\begin{tabular}{c|cc|c|c}
		\multirow{2}{*}{\normalsize{} Method } &  \multicolumn{2}{c|}{\small{} SIFID~$\downarrow$} & \multirow{2}{*}{\small{} LPIPS~$\uparrow$ } & \small{} Dist.~   \tabularnewline
		& \footnotesize{$\frac{H\times W}{4}$} & \footnotesize{$\frac{H\times W}{16}$} & & to train \tabularnewline
		
		\hline 	\hline 	
		
		{\normalsize{} SinGAN} & \color{darkred} \small{2.47}  & \color{darkred} \small{96.35}   &  \small{0.32}  &   \small{0.51}  \tabularnewline

		{\normalsize{} ConSinGAN} & \color{darkred} \small{2.74} & \color{darkred} \small{74.50}  &    \small{0.34}  &   \small{0.53} \tabularnewline
		
		{\normalsize{} FastGAN}  & \small{0.79} & \small{9.24}  & \textbf{\small{0.43}} &  \color{darkred} \small{0.13}  \tabularnewline
		
		{\normalsize{} SIV-GAN }  & \textbf{\small{0.55}} & \textbf{\small{5.14}} & \textbf{\small{0.43}} &  \small{0.34}
		\end{tabular}
\vspace{-1.9em}
\end{table}


%
%
%
%
		
		

		
		

\subsection{Implementation and training} \label{sec: train_det}

The overall architecture of SIV-GAN is shown in Fig. \ref{fig:model_over}. In our implementation, the SIV-GAN generator employs ResNet blocks which are similar to BigGAN \citep{Brock2019}.
However, we do not use BatchNorm or self-attention. 
As in MSG-GAN \citep{karnewar2019msg}, we generate images at intermediate ResNet blocks of $G$, passing them to $D_{low-level}$ to facilitate the gradient flow from the discriminator. The latent vector $z$ of length $64$ is sampled from $N(0,1)$. It is by default broadcasted to the spatial dimensions of 3x5, which can be adjusted to closer fit the shape of a training sample. For diversity regularization, we use the $\mathrm{tanh}$ activation on the features from the final convolutions of the $G$ blocks. 

The SIV-GAN discriminator also uses ResNet blocks. 
We set $N_{low-level} = 3$, $N_{layout} = N_{content} = 4$, thus using $3$ ResNet blocks before branching and $4$ ResNet blocks for the content and layout branches. The impact of the number of blocks will be evaluated in Sec.~\ref{sec:exp_ablations}.
To enable multi-scale gradients, we incorporate images at different scales using the $\phi_{lin\_cat}$ strategy from \cite{karnewar2019msg}. 
The proposed feature augmentation (FA) is applied with probability $0.4$ at every discriminator forward pass.
We also use differentiable image augmentation (DA) \citep{Karras2020TrainingGA,Zhao2020ImageAF,Zhao2020DifferentiableAF}, applying translation, cropping, rotation, and horizontal flipping for real and fake images with a probability of $0.7$ at each forward pass. As in \citep{Karras2020TrainingGA}, we observe no signs of leaking augmentations in the generated samples. 

In contrast to previous single image GANs \citep{Shaham2019SinGANLA,Hinz2020ImprovedTF}, which employ a multi-stage training scheme, SIV-GAN is trained end-to-end in one stage, using the loss from Eq.~\ref{loss_full} with $\lambda$ = $0.15$ for $\mathcal{L}_{DR}$. The impact of $\lambda$ will be evaluated in Sec.~\ref{sec:exp_ablations}. We use spectral normalization~\citep{miyato2018spectral} for both $G$ and $D$, and do not use a reconstruction loss as in~\citep{Shaham2019SinGANLA,Hinz2020ImprovedTF} or any other stabilization techniques. 
SIV-GAN is trained using the ADAM optimizer with $(\beta_1, \beta_2) = (0.5, 0.999)$, a learning rate of $0.0002$ for both $G$ and $D$, and a batch size of $5$ using different augmentations of a single image or video frames.

\section{Experiments}
\label{sec:experiments}

\subsection{Experimental setup}
\label{sec:exp_setup}

We evaluate SIV-GAN by conducting experiments in two different settings: learning from a \textit{single image} and from a \textit{single video}. For both of them, we use the same model configuration as described in Sec.~\ref{sec: train_det}. 
We train our model for 100k iterations in the Single Image setting and for 300k iterations in the Single Video setting.




\begin{figure}[t]
	\vspace{0.8ex}
    \centering
    \includegraphics[width=0.95\linewidth]{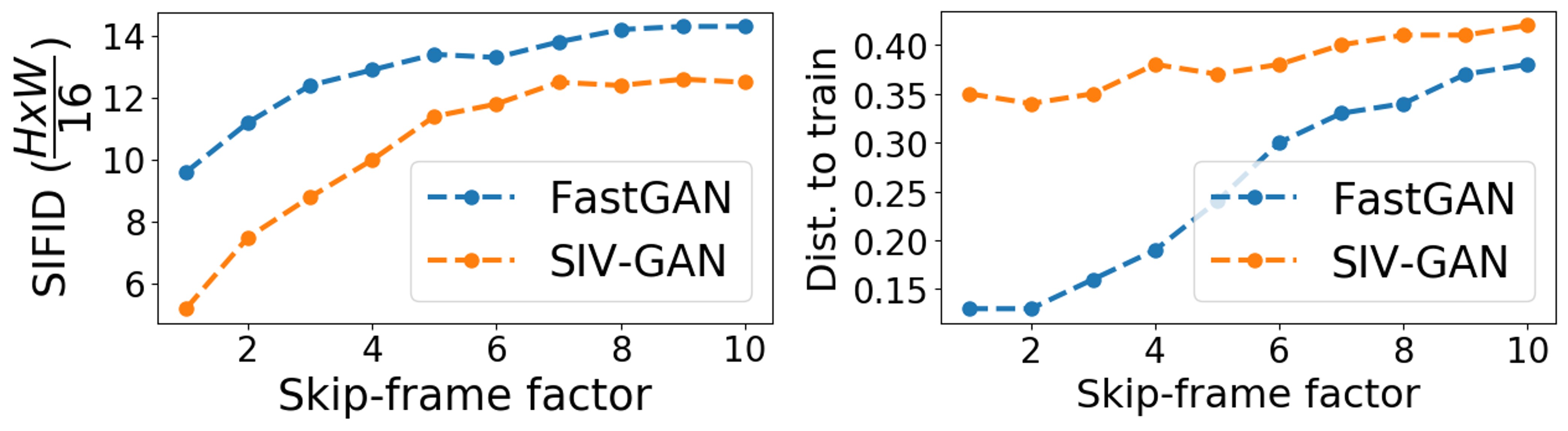}
    \vspace{-1.8ex}
    \caption{\label{fig:correlation} Comparison between SIV-GAN and FastGAN in the Single Video setting. A skip-factor factor of $N$ indicates that every $N$-th frame was sampled for training. A smaller factor means higher average similarity between the sampled training images.  }
    \vspace{-2.3ex}
\end{figure}

\myparagraph{Datasets.} Following SinGAN~\citep{Shaham2019SinGANLA}, we evaluate the Single Image setting on 50 images extracted from the Places dataset \citep{zhou2017places}. In addition to their protocol, we also select 15 videos from the DAVIS~\citep{pont20172017} and YFCC100M \citep{thomee2016yfcc100m} datasets. We use all the frames as training images in the Single Video setting, while we use only one frame from the middle of each video for the Single Image setup. The chosen videos last for 2-10 seconds and consist of 60-100 frames.

\begin{figure*}[t]
\begin{centering}
\setlength{\tabcolsep}{0.1em}
\renewcommand{\arraystretch}{0}
\par\end{centering}
\begin{centering}
\hfill{}%
\begin{tabular}{@{\hskip -0.05in}c@{\hskip 0.04in}c@{\hskip 0.04in}c@{\hskip 0.04in}c@{\hskip 0.04in}|@{\hskip 0.04in}c@{\hskip 0.04in}c@{\hskip 0.04in}c@{\hskip 0.04in}c}

\rotatebox{90}{ \hspace{1.7ex}\begin{tabular}{c} Video \\ frames\end{tabular}} &  \multicolumn{3}{c@{\hskip 0.04in}|@{\hskip 0.04in}}{\hspace{-2.05ex} \includegraphics[width=0.463\linewidth, height=0.075\textheight]{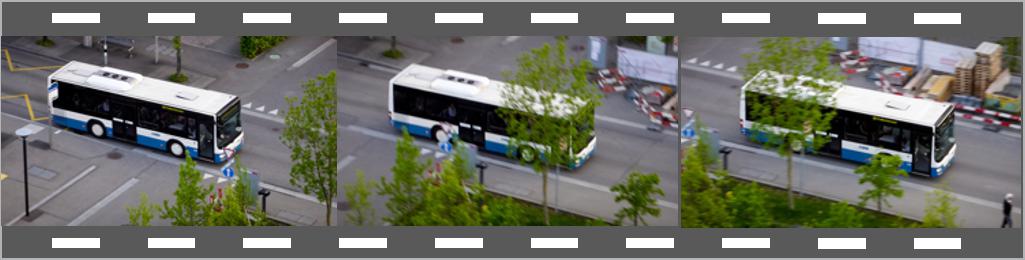} } &
\multicolumn{3}{@{\hskip -0.00in}c@{\hskip 0.04in}}{\includegraphics[width=0.463\linewidth, height=0.075\textheight]{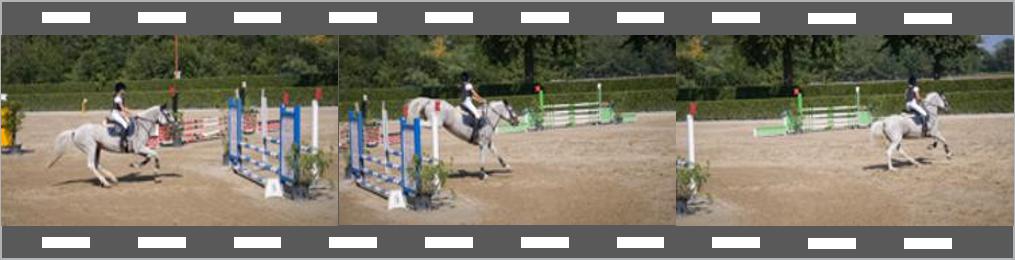}} \tabularnewline[-7pt]
\rotatebox{90}{\hspace{-.3ex}\begin{tabular}{c} ~SinGAN \end{tabular}}&
\includegraphics[width=0.15\linewidth, height=0.062\textheight]{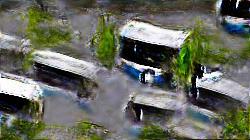} & 
\includegraphics[width=0.15\linewidth, height=0.062\textheight]{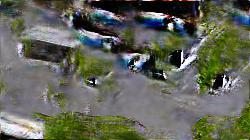} & 
\includegraphics[width=0.15\linewidth, height=0.062\textheight]{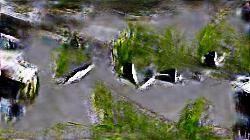} &
\includegraphics[width=0.15\linewidth, height=0.062\textheight]{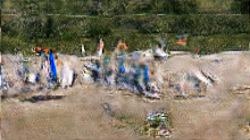} & 
\includegraphics[width=0.15\linewidth, height=0.062\textheight]{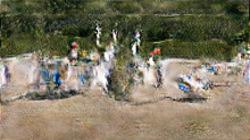} &
\includegraphics[width=0.15\linewidth, height=0.062\textheight]{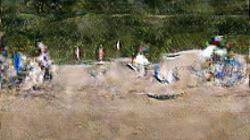}
\tabularnewline[-6pt]

\rotatebox{90}{\hspace{-.6ex}\begin{tabular}{c} FastGAN  \end{tabular}}&
\includegraphics[width=0.15\linewidth, height=0.062\textheight]{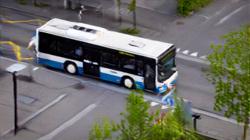} & 
\includegraphics[width=0.15\linewidth, height=0.062\textheight]{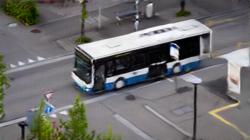} & 
\includegraphics[width=0.15\linewidth, height=0.062\textheight]{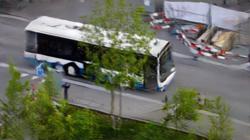} & 
\includegraphics[width=0.15\linewidth, height=0.062\textheight]{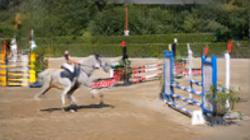} &
\includegraphics[width=0.15\linewidth, height=0.062\textheight]{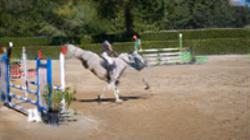} & 
\includegraphics[width=0.15\linewidth, height=0.062\textheight]{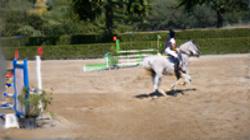}
\tabularnewline[-8pt]

\rotatebox{90}{\hspace{-.5ex}\begin{tabular}{c}  SIV-GAN \end{tabular}}&
\includegraphics[width=0.15\linewidth, height=0.062\textheight]{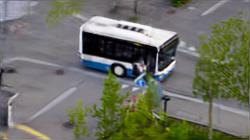} &
\includegraphics[width=0.15\linewidth, height=0.062\textheight]{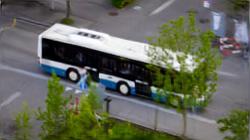} & 
\includegraphics[width=0.15\linewidth, height=0.062\textheight]{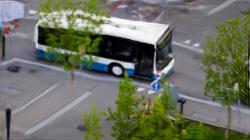} & 
\includegraphics[width=0.15\linewidth, height=0.062\textheight]{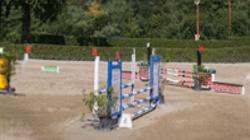} &
\includegraphics[width=0.15\linewidth, height=0.062\textheight]{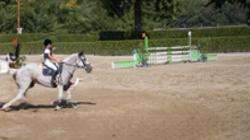} & 
\includegraphics[width=0.15\linewidth, height=0.062\textheight]{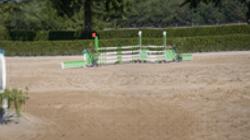}

\end{tabular}\hfill{}
\par\end{centering}
\vspace{-0.75em}
\caption{\label{fig:qual_video} Visual comparison in the Single Video setting. While other models only reproduce the training frames or fail to correctly generate objects, SIV-GAN produces high-quality images substantially different from the original training frames.
 }
\end{figure*}

\myparagraph{Metrics.} To assess the quality of generated images, we measure the mean single FID (SIFID) \citep{Shaham2019SinGANLA}. Following the evaluation from \cite{Shaham2019SinGANLA} and  \cite{Hinz2020ImprovedTF}, in the Single Image setting we also report the \textit{best} SIFID among the generated samples. The original SIFID formulation uses InceptionV3 features before the first pooling layer at $\frac{H\times W}{4}$ resolution. We observed that such metric captures only low-level image details, such as colors and textures, and not high-level semantic image properties, such as appearance of objects or global layouts. To evaluate higher-level realism,
we therefore additionally use later features obtained before the final classification layer at $\frac{H\times W}{16}$ resolution. 
To evaluate the diversity of samples, we adopt the pixel diversity metric from \cite{Shaham2019SinGANLA}. To measure perceptual diversity,
we also report the average LPIPS \citep{Alexey2016} across pairs of generated images. To verify that the models do not simply reproduce the training set, we report average LPIPS to the nearest image in the training set, augmented in the same way as during training (Dist.\ to train).
We note that 
SIFID tends to penalize diversity, favouring overfitting \citep{Robb2020FewShotAO}. To account for this quality-diversity trade-off, a fair analysis should thus assess both diversity and quality.

\myparagraph{Comparison models.} We compare our model with two single-image GANs, SinGAN \citep{Shaham2019SinGANLA} and ConSinGAN \citep{Hinz2020ImprovedTF}, and with a GAN model for few-shot image synthesis, FastGAN \citep{anonymous2021towards}. We use the original implementation source codes provided by the authors. While training single image GANs \citep{Shaham2019SinGANLA,Hinz2020ImprovedTF} on a single video, we applied the reconstruction loss on all frames, as we found this helpful in stabilizing the training. In addition, we compare our model to DreamBooth, a pre-trained text-to-image diffusion model that can be fine-tuned using a single image or few video frames. We use a publicly available re-implementation\footnote{\url{https://github.com/XavierXiao/Dreambooth-Stable-Diffusion}} of this model.
The source code of SIV-GAN is publicly available\footnote{\url{https://github.com/boschresearch/one-shot-synthesis}} at a repository which also contains the source code of the work \citep{sushko2023wacv}.

\subsection{Comparison to previous GAN models}
\label{sec:exp_sin_vid}

Tables~\ref{table:comp_single_image} and \ref{table:comp_single_video} present a quantitative comparison between the models in the Single Image and Single Video settings, while the respective visual results are shown in Figs.~\ref{fig:qual_image} and \ref{fig:qual_video}. As seen from the tables, SIV-GAN notably outperforms other models in both studied settings. Despite a potential trade-off between quality and diversity, our model achieves better performance in both, reaching lower SIFID values and higher diversity scores. Importantly, only SIV-GAN successfully learns from both single images and videos, generating globally-coherent images of high diversity. Next, we analyse results in these settings separately.

\begin{figure*}[t]
	\begin{centering}
		\setlength{\tabcolsep}{0.0em}
		\renewcommand{\arraystretch}{0}
		\par\end{centering}
	\begin{tabular}{c@{\hskip 0.10in}c@{\hskip 0.05in}c@{\hskip 0.05in}c@{\hskip 0.05in}c@{\hskip 0.05in}}
			
			Training image & \multicolumn{4}{c}{Generated samples }
			\tabularnewline 
			
			\includegraphics[width=0.19\linewidth,height=0.115\linewidth]{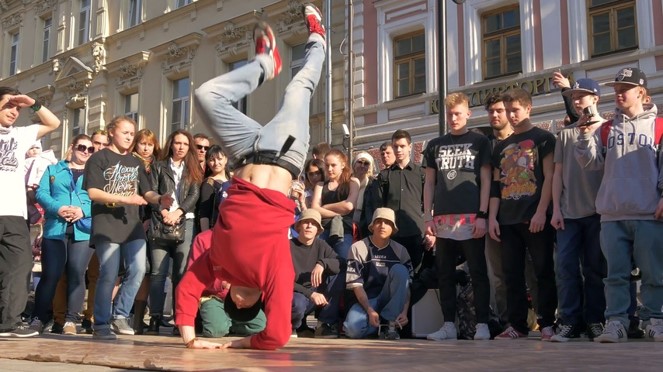}   & 	

			\includegraphics[width=0.19\linewidth,height=0.115\linewidth]{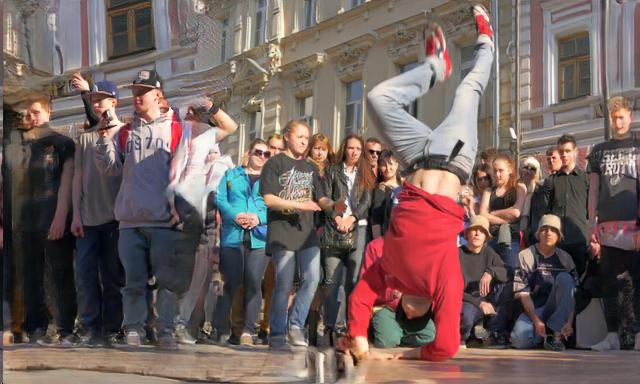} &
			\includegraphics[width=0.19\linewidth,height=0.115\linewidth]{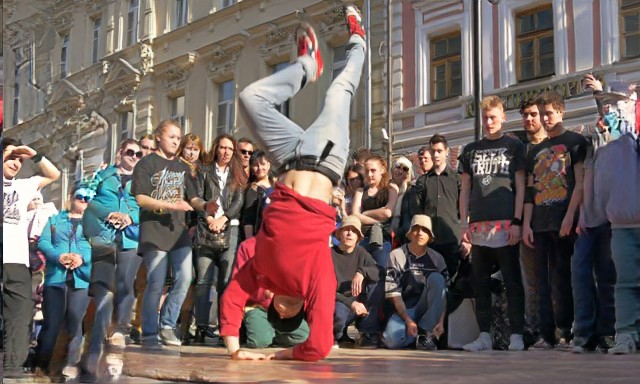} & 
			\includegraphics[width=0.19\linewidth,height=0.115\linewidth]{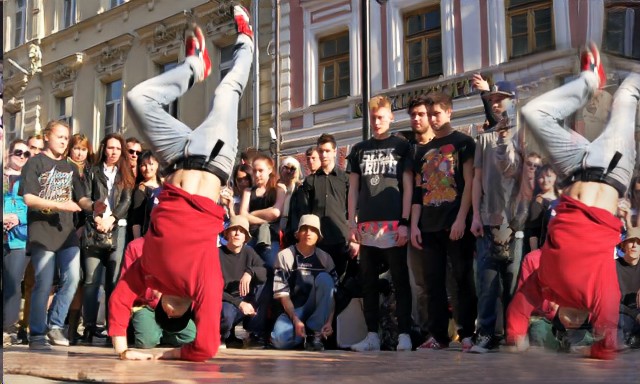} &
			\includegraphics[width=0.19\linewidth,height=0.115\linewidth]{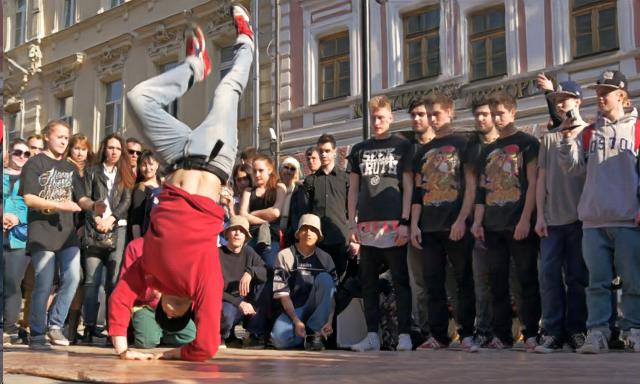} 
\tabularnewline 

            \includegraphics[width=0.19\linewidth,height=0.115\linewidth]{supplementary/figures/higher_res/ref5.jpg}   & 	

			\includegraphics[width=0.19\linewidth,height=0.115\linewidth]{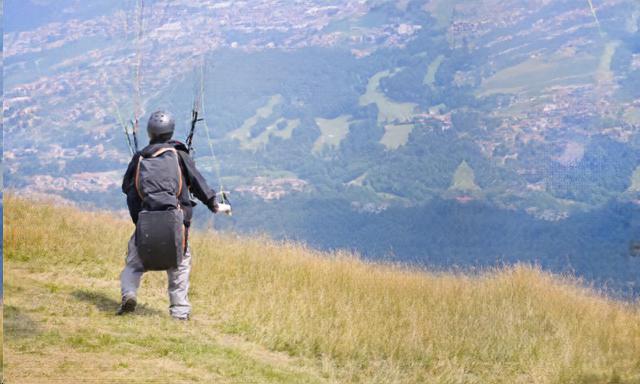} &
			\includegraphics[width=0.19\linewidth,height=0.115\linewidth]{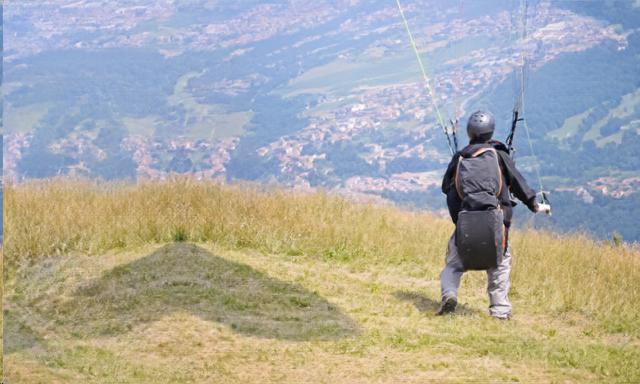} & 
			\includegraphics[width=0.19\linewidth,height=0.115\linewidth]{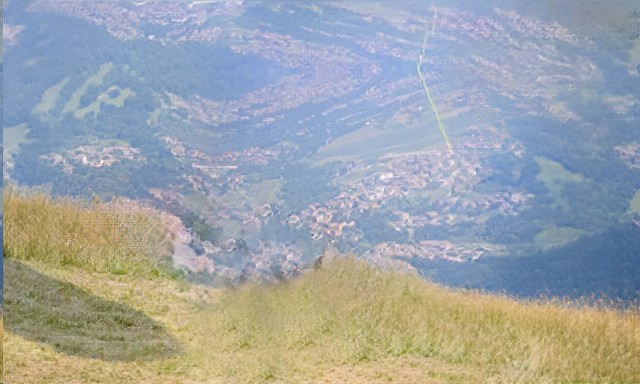} &
			\includegraphics[width=0.19\linewidth,height=0.115\linewidth]{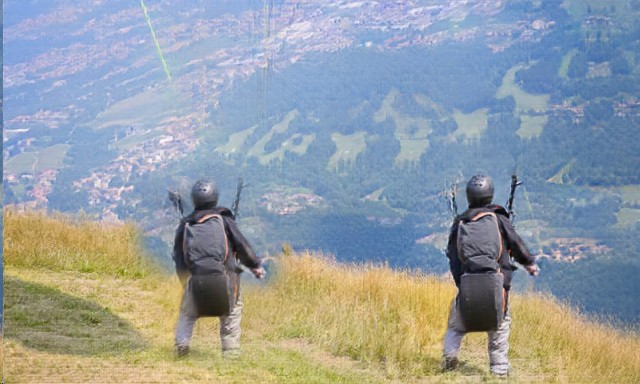} 
\tabularnewline 

            \includegraphics[width=0.19\linewidth,height=0.115\linewidth]{supplementary/figures/higher_res/ref6.jpg}   & 	

			\includegraphics[width=0.19\linewidth,height=0.115\linewidth]{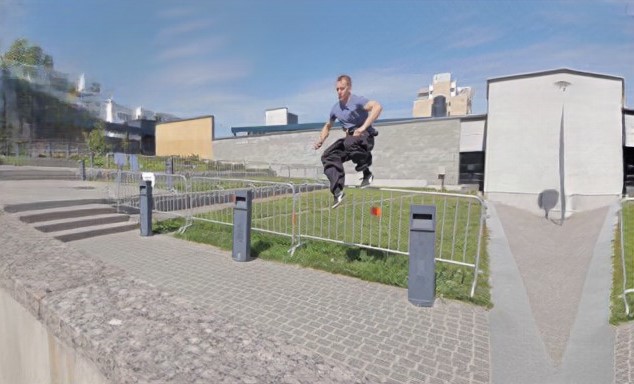} &
			\includegraphics[width=0.19\linewidth,height=0.115\linewidth]{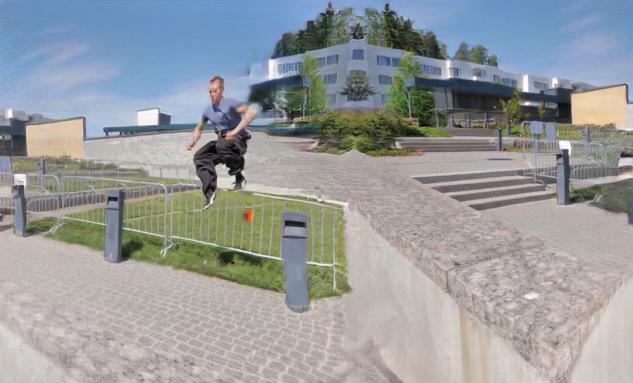} & 
			\includegraphics[width=0.19\linewidth,height=0.115\linewidth]{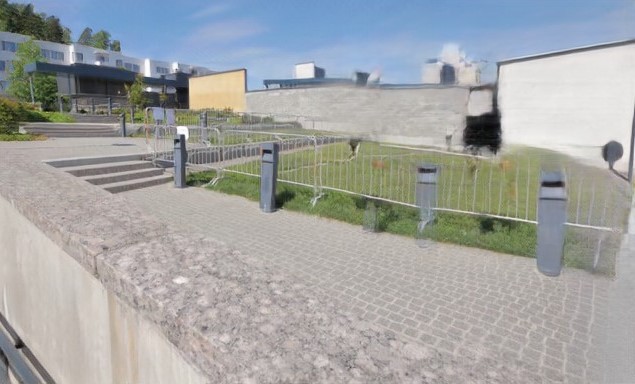} &
			\includegraphics[width=0.19\linewidth,height=0.115\linewidth]{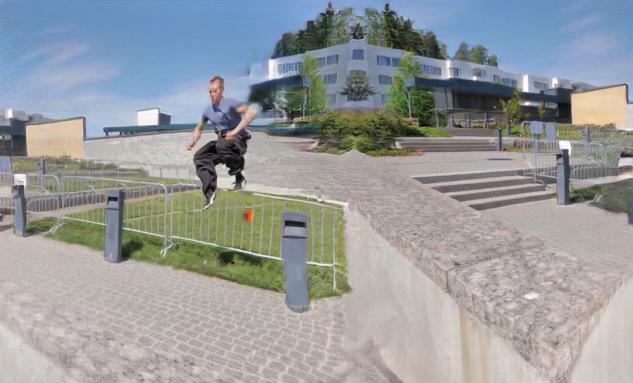}

		\end{tabular}

	\caption{Additional results of SIV-GAN in the Single Image setting. Given a single image for training, our model can produce novel scene compositions of the same scene, modifying the layouts of background, changing the number of foreground objects, or changing their positions in the scene.}
	\label{fig:high_res_cviu}
\end{figure*}

\begin{figure*}[h!]
\begin{centering}
\setlength{\tabcolsep}{0.0em}
\renewcommand{\arraystretch}{1.5}
\par\end{centering}
\begin{centering}
\centering
\begin{tabular}{c@{\hskip 0.05in}c@{\hskip 0.05in}c@{\hskip 0.05in}c@{\hskip 0.05in}c@{\hskip 0.05in}c}

\multicolumn{6}{c}{
	\includegraphics[width=0.95\linewidth]{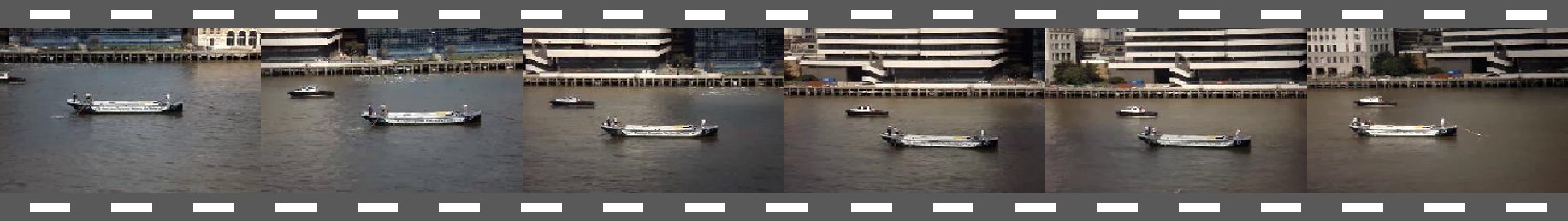}
}

\tabularnewline

\includegraphics[width=0.15255\linewidth]{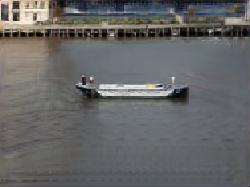} & 
\includegraphics[width=0.1525\linewidth]{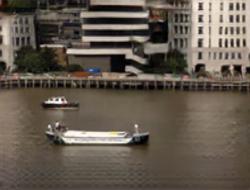} & 
\includegraphics[width=0.1525\linewidth]{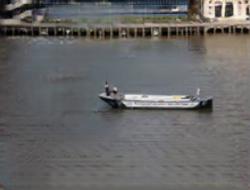} & 
\includegraphics[width=0.1525\linewidth]{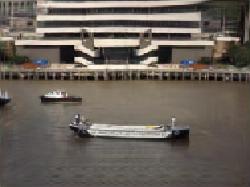} & 
\includegraphics[width=0.1525\linewidth]{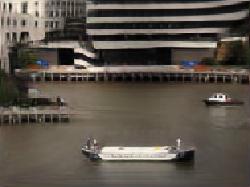} & 
\includegraphics[width=0.1525\linewidth]{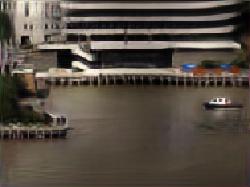} \tabularnewline 

\multicolumn{6}{c}{
	\includegraphics[width=0.95\linewidth]{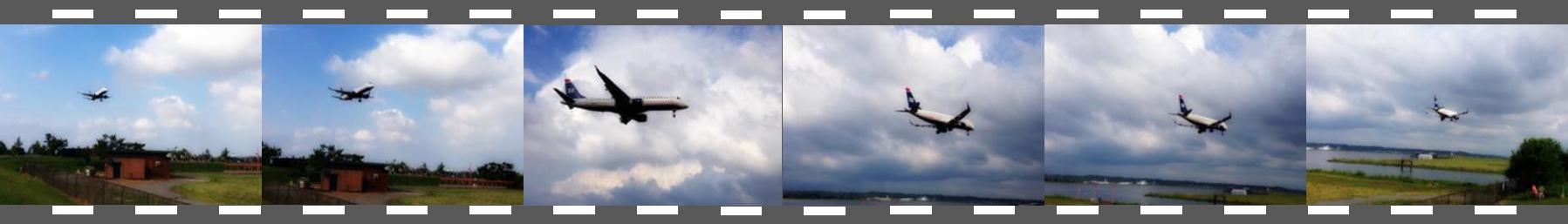}
}

\tabularnewline

\includegraphics[width=0.1525\linewidth]{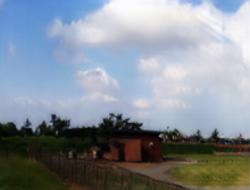} & 
\includegraphics[width=0.1525\linewidth]{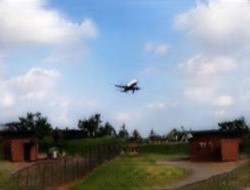} & 
\includegraphics[width=0.1525\linewidth]{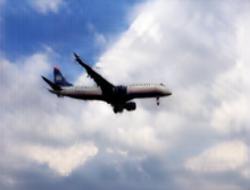} & 
\includegraphics[width=0.1525\linewidth]{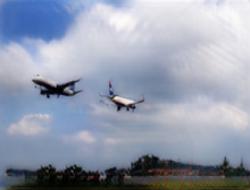} & 
\includegraphics[width=0.1525\linewidth]{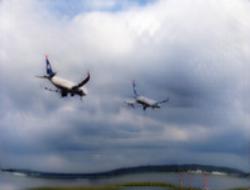} & 
\includegraphics[width=0.1525\linewidth]{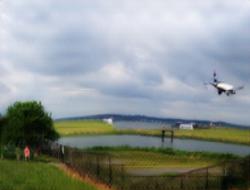} \tabularnewline \tabularnewline

\end{tabular}
\par\end{centering}
\caption{\label{fig:qual_video_app_cviu} Additional qualitative results in the Single Video setting. The training sequences are shown in grey frames. Given a single video for training, our model produces images that are different to the training frames. The generated images are shown next to their closest training frames by LPIPS.}
\end{figure*}

\myparagraph{Single Image.}
\label{sec:exp_sin_image}
Given a single image for training, SIV-GAN produces diverse samples of high visual quality as shown in Figs.~\ref{fig:teaser} and \ref{fig:qual_image}. For the example in Fig.~\ref{fig:qual_image}, our model can change the number and placement of foreground objects (e.g., fish and people), or edit the contour and position of rocks in landscape images. 
Note that such changes preserve the original scene context, retaining the appearance of objects and maintaining the scene layout realism. In contrast, the prior single image GAN models, SinGAN and ConSinGAN, tend to disturb the appearance of objects (e.g., by washing away the contours of fish and people) and disrespect layouts (e.g., sky textures can appear below the horizon), while exhibiting lower diversity in content and layouts.
This is reflected in their higher SIFID and lower diversity scores in Table~\ref{table:comp_single_image}. On the other hand, the few-shot FastGAN model suffers from memorization issues, only reproducing the training image or its flipped version, which was generated during training by data augmentation. This is reflected in Table~\ref{table:comp_single_image} by the lowest diversity and Dist.\ to train (in red) metrics on both datasets. Despite having the lowest diversity, we observe that FastGAN does not reach a low SIFID due to leaking augmentations (horizontal flipping).

\begin{figure*}[t]
\begin{centering}
\setlength{\tabcolsep}{0.1em}
\renewcommand{\arraystretch}{0}
\par\end{centering}
\begin{centering}
\hfill{}%
\begin{tabular}{@{\hskip -0.05in}c@{\hskip 0.04in}c@{\hskip 0.04in}c@{\hskip 0.04in}c@{\hskip 0.04in}|@{\hskip 0.04in}c@{\hskip 0.04in}c@{\hskip 0.04in}c@{\hskip 0.04in}c}

\rotatebox{90}{ \hspace{1.7ex}\begin{tabular}{c} Train. \\ frames\end{tabular}} &  \multicolumn{3}{c@{\hskip 0.04in}|@{\hskip 0.04in}}{\hspace{-2.05ex} \includegraphics[width=0.15\linewidth, height=0.075\textheight]{figures/bank_sin_im/fish/ref} } &
\multicolumn{3}{@{\hskip -0.00in}c@{\hskip 0.04in}}{\includegraphics[width=0.463\linewidth, height=0.075\textheight]{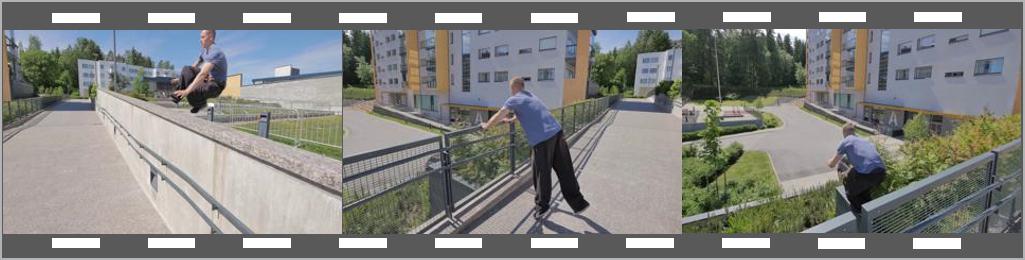}} \tabularnewline[-11pt]

\rotatebox{90}{\hspace{-.3ex}\begin{tabular}{c} \hspace{-1.5ex} \small{DreamBooth} \end{tabular}}&
\includegraphics[width=0.15\linewidth, height=0.062\textheight]{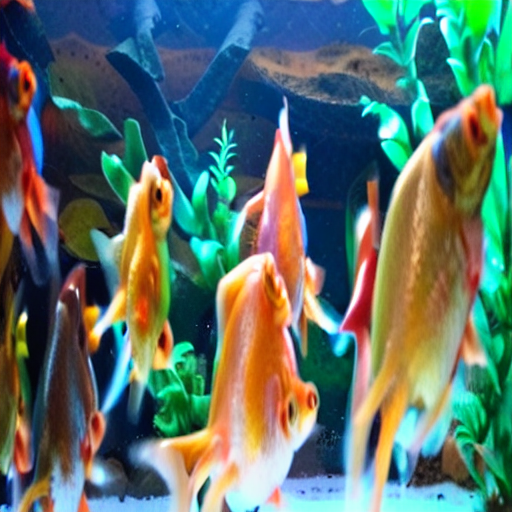} & 
\includegraphics[width=0.15\linewidth, height=0.062\textheight]{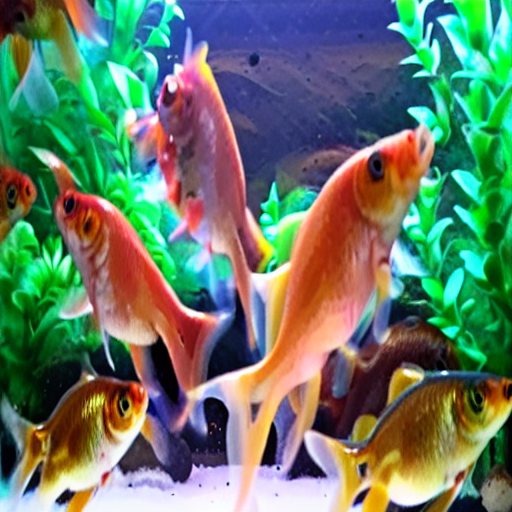} & 
\includegraphics[width=0.15\linewidth, height=0.062\textheight]{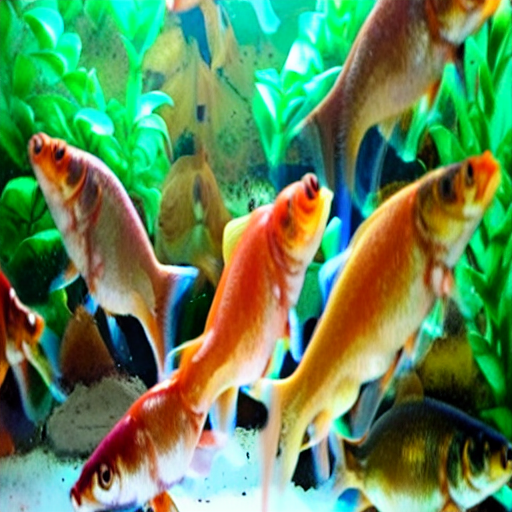} &
\includegraphics[width=0.15\linewidth, height=0.062\textheight]{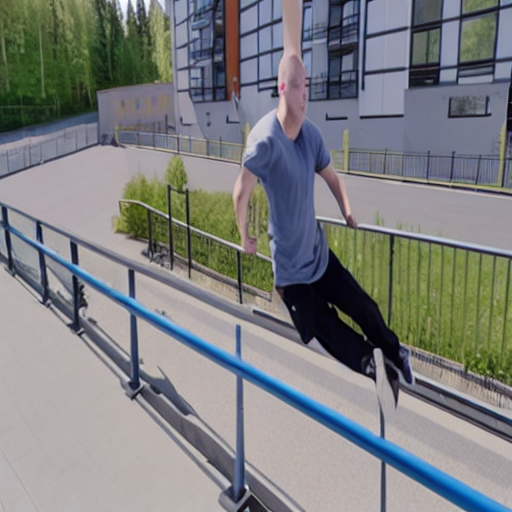} & 
\includegraphics[width=0.15\linewidth, height=0.062\textheight]{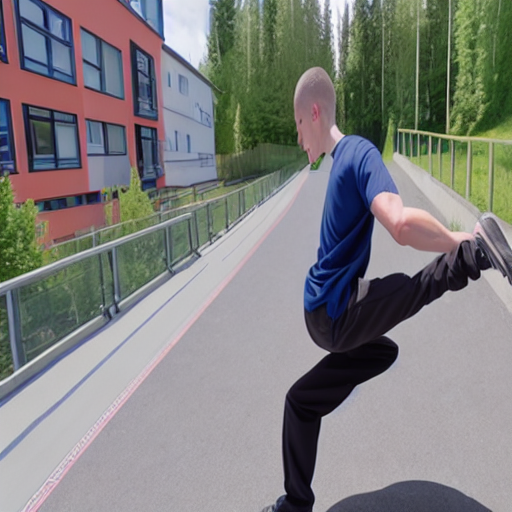} &
\includegraphics[width=0.15\linewidth, height=0.062\textheight]{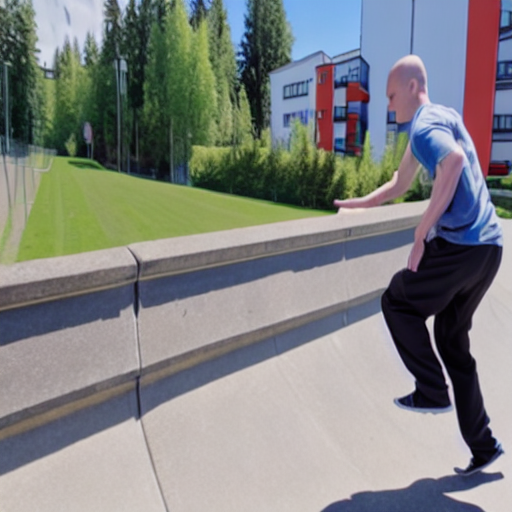}
\tabularnewline[-6pt]

\rotatebox{90}{\hspace{-.5ex}\begin{tabular}{c} \hspace{-1.0ex} SIV-GAN \end{tabular}}&
\includegraphics[width=0.15\linewidth, height=0.062\textheight]{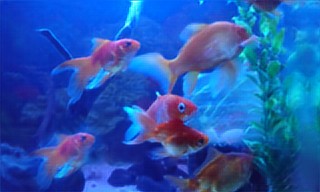} &
\includegraphics[width=0.15\linewidth, height=0.062\textheight]{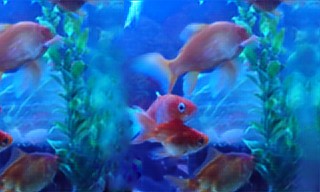} & 
\includegraphics[width=0.15\linewidth, height=0.062\textheight]{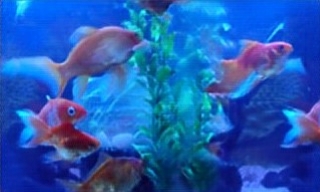} & 
\includegraphics[width=0.15\linewidth, height=0.062\textheight]{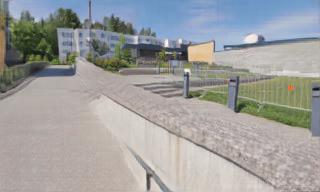} &
\includegraphics[width=0.15\linewidth, height=0.062\textheight]{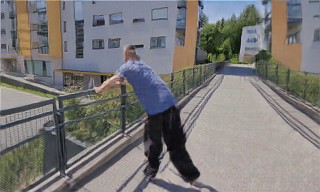} & 
\includegraphics[width=0.15\linewidth, height=0.062\textheight]{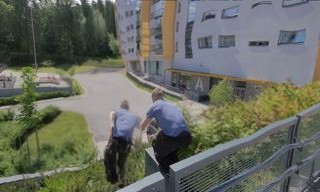}

\end{tabular}\hfill{}
\par\end{centering}
\vspace{-0.75em}
\caption{\label{fig:comp_dreambooth} Visual comparison to DreamBooth \citep{ruiz2023dreambooth} in the Single Image and Video settings. As DreamBooth was pre-trained on a large-scale dataset, it is not inherently limited by the appearance of objects present in original training samples, which allows achieving higher diversity. On the other hand, the fine-tuned DreamBooth is prone to errors, e.g., failing to tune to the distribution of colors to a new image correctly or distorting objects, for example, generating fish with two heads or humans in unrealistic poses. }
\end{figure*}

\myparagraph{Single Video.} 
The Single Video setting provides multiple video frames for training. Consequently, generative models have the potential to combine the knowledge observed in different video frames, and thereby to synthesize more interesting combinations of objects and scenes. Figs.~\ref{fig:teaser} and \ref{fig:qual_video} show the images generated by SIV-GAN in this setting. Our model generates high-quality images that are substantially different from the training frames, adding/removing objects and changing the scene geometry.
For example, having seen a car following a road (Fig.~\ref{fig:teaser}), SIV-GAN generates the scene without a car or with two cars. In Fig.~\ref{fig:qual_video}, our model varies the length of a bus and placement of trees, or removes a horse from the scene and changes the jumping obstacle configuration. 
In contrast, SinGAN, which has been developed to learn from a single image, does not generalize to the Single Video setting.  
It distorts objects and produces unrealistic layouts, resulting in an extremely high SIFID in Table \ref{table:comp_single_video}. The few-shot FastGAN, on the other hand, generates images with reasonable fidelity, but is still unable to produce samples with non-trivial layout changes,   
achieving only a very low Dist.\ to train score ($0.13$ in Table \ref{table:comp_single_video}). We conclude that only SIV-GAN deals with the challenging Single Video setting successfully, producing globally-coherent images and avoiding memorization of training data.

\begin{table}[t]
	\setlength{\tabcolsep}{0.4em}
	\renewcommand{\arraystretch}{1.15}
	\centering
	\caption{Comparison of synthesis quality and diversity between SIV-GAN and DreamBooth \citep{ruiz2023dreambooth} on DAVIS-YFCC100M. * indicates that the model requires textual inputs describing training images or videos. }

	\begin{tabular}{c|cc|cc}

	\normalsize \multirow{2}{*}{\small Model}  & \multicolumn{2}{c|}{Single Image} & \multicolumn{2}{c}{Single Video}  \tabularnewline

	\normalsize  & {\small SIFID~$\downarrow$}  & {\small LPIPS~$\uparrow$ } & {\small SIFID~$\downarrow$}  & {\small LPIPS~$\uparrow$ }  \tabularnewline

		\hline 	\hline 	
		
		 \normalsize SIV-GAN & \small \textbf{{0.08}}  &  \small {{0.33}}  & \small \textbf{{0.55}}  &  \small {{0.43}} \tabularnewline
		 \normalsize DreamBooth* & \small {{0.15}}  &  \small \textbf{{0.42}}  & \small {{0.98}}  &  \small \textbf{{0.54}}

		\end{tabular}
\label{tab:comp_dreambooth} %

\end{table}

In Fig.~\ref{fig:correlation} we provide an extended analysis, exploring the performance of SIV-GAN and FastGAN while changing the sampling rate.  
For this, we take 5 long video sequences from YFCC100M \citep{thomee2016yfcc100m} and construct 10 different subsets for each of them: a subset with index $i$ includes every $i^{th}$ video frame. The number of sampled frames is for all subsets 100. This way, a lower skip-frame factor indicates that the chosen frames are closer in time and thus more similar to each other. 
As seen from Fig.~\ref{fig:correlation}, FastGAN suffers from memorization the most when learning from a set of very similar images: its Dist.\ to train scores fall dramatically when the skip-frame factor is decreased. This indicates that learning from a few-shot dataset consisting of very similar images can be challenging for prior few-shot GAN models. In contrast, SIV-GAN preserves high Dist.\ to train scores even for lowest skip-frame factors. Notably, our model also outperforms FastGAN in SIFID for all skip-frame factors.

\myparagraph{Additional qualitative results.}
In Fig.~\ref{fig:high_res_cviu} and \ref{fig:qual_video_app_cviu}, we provide additional qualitative results of SIV-GAN in both the Single Image and Single Video settings. 
In Fig.~\ref{fig:high_res_cviu}, given a single image showing a dancer, paraglider, or parkour jumper, SIV-GAN re-arranges the scene by placing people in new locations and composing new layouts for backgrounds. Importantly, the boundaries of objects are not distorted and the scenes remain realistic at a global scale. Similarly, while learning from multiple frames of videos in Fig.~\ref{fig:qual_video_app_cviu}, SIV-GAN produces novel scene compositions by combining objects from different frames (e.g., boat shown in a new place or plane shown from a different angle).

\subsection{Comparison to fine-tuned diffusion model DreamBooth}

In addition to single-image and few-shot GAN models, we provide a comparison to a more recent diffusion model DreamBooth \citep{ruiz2023dreambooth}. This model is based on Latent Diffusion \citep{rombach2022high}, a text-to-image diffusion model that was pre-trained on $\sim$400M text-image pairs, but can be fine-tuned in small data regimes like on a single image or several frames of a single video. To apply DreamBooth in these regimes, we manually input the model with textual descriptions of objects present in training images or videos.

The comparison results are shown in Fig.~\ref{fig:comp_dreambooth} and Table~\ref{tab:comp_dreambooth}. As seen in Fig.~\ref{fig:comp_dreambooth}, DreamBooth achieves a similar ability for generating new scene compositions based on limited data. For example, DreamBooth generates new fish in different locations from a single image or a parkour jumper in new environments when fine-tuned on frames of a single video. We observe that DreamBooth generally achieves higher diversity of synthesis, as it has no inherent limitation to generate only the objects that were seen in the one or few training samples. Such higher diversity is generally expected, as, in contrast to SIV-GAN that is trained from scratch, DreamBooth was pre-trained on a large-scale dataset containing several hundreds of millions of image-text pairs. However, we also noted that pre-training frequently leads DreamBooth to generating images that deviate from the original samples in colors and textures (e.g., higher saturation). For example, in Fig.~\ref{fig:comp_dreambooth} (left), fine-tuning of DreamBooth converged to the generated images of fish that have a significantly different color palette. 
In addition, higher diversity of DreamBooth commonly leads it to distortions in objects, such as fish with two heads or humans in unrealistic poses in Fig.~\ref{fig:comp_dreambooth}. 
We note that in many applications such artifacts are undesirable, since they corrupt objects' identities and lead to images that are outside of the training distribution. 
Lastly, our observations are reflected in the quantitative comparison presented in Table~\ref{tab:comp_dreambooth}. In both the Single Image and Video settings, SIV-GAN achieves better SIFID scores but lags behind DreamBooth in LPIPS.

\begin{figure*}
\begin{centering}
\setlength{\tabcolsep}{0.0em}
\renewcommand{\arraystretch}{0}
\par\end{centering}
\begin{centering}
\begin{tabular}{@{\hskip -0.03in}c@{\hskip 0.05in}c@{\hskip 0.05in}c@{\hskip 0.05in}c@{\hskip 0.05in}c}
& {No branches} & {No layout branch}  & {No content branch}  & {Both branches}
\tabularnewline	
  \multirow{2}{*}{\begin{tabular}{c} \vspace{-8ex} \\ Training image \\ 	\includegraphics[width=0.18\linewidth, height=0.080\textheight]{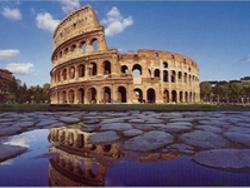} \end{tabular} }&

\includegraphics[width=0.18\linewidth, height=0.080\textheight]{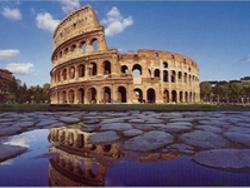} & 
\includegraphics[width=0.18\linewidth, height=0.080\textheight]{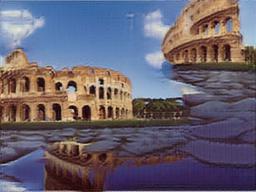} & 
\includegraphics[width=0.18\linewidth, height=0.080\textheight]{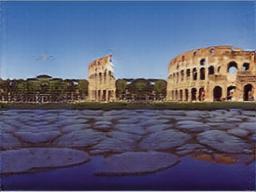} & 
\includegraphics[width=0.18\linewidth, height=0.080\textheight]{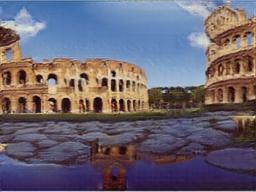} 

\tabularnewline 

& 

\includegraphics[width=0.18\linewidth, height=0.080\textheight]{supplementary/figures/qual_ablations_d/col_no_br1r} & 
\includegraphics[width=0.18\linewidth, height=0.080\textheight]{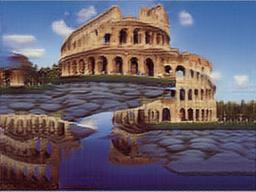} & 
\includegraphics[width=0.18\linewidth, height=0.080\textheight]{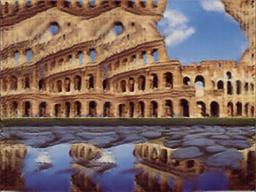} & 
\includegraphics[width=0.18\linewidth, height=0.080\textheight]{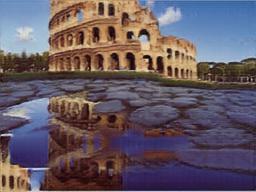}\vspace{0.9ex} \tabularnewline 



\multirow{2}{*}{\begin{tabular}{c} \vspace{-8ex} \\ Training image \\\includegraphics[width=0.18\linewidth, height=0.080\textheight]{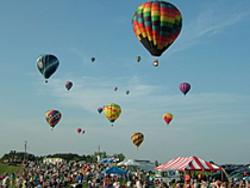} \end{tabular}} &

\includegraphics[width=0.18\linewidth, height=0.080\textheight]{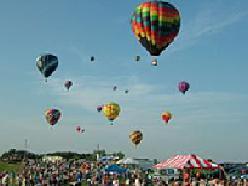} & 
\includegraphics[width=0.18\linewidth, height=0.080\textheight]{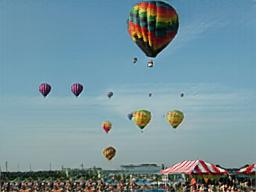} & 
\includegraphics[width=0.18\linewidth, height=0.080\textheight]{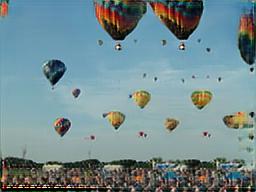} & 
\includegraphics[width=0.18\linewidth, height=0.080\textheight]{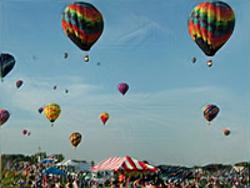} 

\tabularnewline
&
\includegraphics[width=0.18\linewidth, height=0.080\textheight]{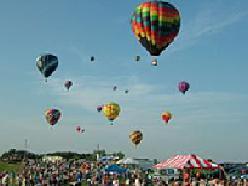} & 
\includegraphics[width=0.18\linewidth, height=0.080\textheight]{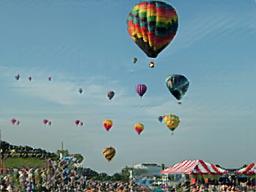} & 
\includegraphics[width=0.18\linewidth, height=0.080\textheight]{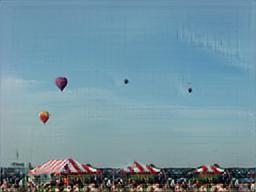} & 
\includegraphics[width=0.18\linewidth, height=0.080\textheight]{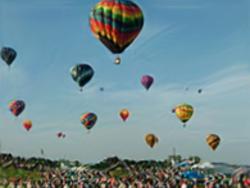} 


\end{tabular}
\par\end{centering}
\vspace{-0.5em}
\caption{\label{fig:qual_ablations_d} Visual results for the ablation on the two-branch discriminator in the Single Image setting. The model with a standard GAN discriminator (no branches) memorizes the training image. The model without the layout branch fails to produce images with realistic layouts or positioning of objects. Absence of the content branch leads to a model that does not preserve well the appearance of objects. Finally, the model with both branches generates diverse images with realistic content and layouts. The quantitative comparison of these models is presented in Table \ref{table:main_ablation}.}
\end{figure*}
\begin{table*}[t]
	\vspace{-0.5ex}
	\setlength{\tabcolsep}{0.4em}
	\renewcommand{\arraystretch}{1.00}
	\centering
	
	\caption{Ablation study in the Single Image and Video settings on DAVIS-YFCC100M. Indicators of collapsed diversity (low LPIPS, Pixel Diversity) or poor quality (high SIFID) are marked in red.}
	\begin{tabular}{c|cc|c|c|c||cc|c|c}
		& \multicolumn{5}{c||}{Single Image}  & \multicolumn{4}{c}{Single Video}  
		\tabularnewline
		\multirow{2}{*}{\normalsize{} Method } & \multicolumn{2}{c|}{\small{} SIFID~$\downarrow$}  & \multirow{2}{*}{\small{} LPIPS~$\uparrow$ } & {\small{} Pixel~$\uparrow$ } & {\small{} Dist.} &  \multicolumn{2}{c|}{\small{} SIFID~$\downarrow$} & \multirow{2}{*}{\small{} LPIPS~$\uparrow$ } &  \small{} Dist.~ \tabularnewline
		& \footnotesize{}  \footnotesize{$\frac{H\times W}{4}$} & \footnotesize{}  \footnotesize{$\frac{H\times W}{16}$} & & \small{} diversity & {\small{} to train} &\footnotesize{$\frac{H\times W}{4}$} & \footnotesize{$\frac{H\times W}{16}$} & & to train  \tabularnewline
		
		\hline 	\hline 	
		{\normalsize{} Full model   } & \small{0.08} & \small{16.30} &  \small{0.33} & \small{0.66} & \small{0.37} &  \small{0.55} & \small{} 5.14 & \textbf{\small{0.43}}  & \small{0.34} 
		 \tabularnewline

		 	\hline
		 {\normalsize{} No Layout br.   } & \color{darkred} \small{0.14}   & \color{darkred} \small{20.29}  & \textbf{\small{0.35}} & \textbf{\small{0.67}} & \small{0.40}  & \color{darkred} \small{} 0.71 & \color{darkred} \small{11.70}   & \small{0.42}  & \small{0.38}   \tabularnewline
		 
		 {\normalsize{} No Content br.   } & \small{0.08} & \color{darkred}\small{23.25}  &  \small{0.34} &  \small{0.64} & \small{0.36} & \color{darkred} \small{0.73}  & \color{darkred} \small{10.43}   & \small{0.41} & \small{0.33}   \tabularnewline
		 
		 {\normalsize{} No branches  } & \textbf{\textbf{\small{0.03}}} & \textbf{\textbf{\small{7.73}}}  &  \color{darkred}\small{0.13} & \color{darkred} \small{0.43} & \color{darkred}  \small{0.12} & \small{0.42}  & \textbf{\small{3.73}}  & \small{} \color{darkred} 0.37 & \color{darkred} \small{0.18}  
		 \tabularnewline

		 \hline

		{\normalsize{} No DR } & \small{0.05} & \small{11.99}  &  \color{darkred} \small{0.04} & \color{darkred} \small{0.33} & \color{darkred}  \small{0.06} & \textbf{\small{0.40}}  & \small{9.81}  & \color{darkred} \small{0.30}   &  \small{0.32}    \tabularnewline 
		
		{\normalsize{} No FA } & \small{0.08} & \small{14.81}  &  \small{0.27} &   \small{0.58} &  \small{0.33} & \small{0.51}  & \small{4.85} & \small{0.41}  &  \small{0.32}   \tabularnewline 
			 {\normalsize{} No $\mathcal{L}_{D_{low-level}}$ } & \small{0.08} & \small{15.92}  &  \small{0.27} & \small{0.56} & \small{0.29} & \small{0.58}  & \small{5.32}  & \small{0.40}  & \small{0.31}   \tabularnewline
		\end{tabular}
\label{table:main_ablation} %

\end{table*}

\subsection{Ablations}
\label{sec:exp_ablations}

\myparagraph{Ablations on the main model components.}
In Table \ref{table:main_ablation}, we demonstrate the importance of the main components of SIV-GAN. In each row, we remove only one model component, starting from the full SIV-GAN model. 

\begin{figure*}[h!]
\begin{centering}
\setlength{\tabcolsep}{0.0em}
\renewcommand{\arraystretch}{0}
\par\end{centering}
\begin{centering}
\vspace{-0.4em}
\begin{tabular}{@{\hskip -0.03in}c@{\hskip 0.0in}c@{\hskip 0.05in}c@{\hskip 0.11in}c@{\hskip 0.05in}c@{\hskip 0.11in}c@{\hskip 0.05in}c@{}}
 & \multicolumn{2}{c}{SIV-GAN without DR} &\multicolumn{2}{c}{SIV-GAN without FA}  & \multicolumn{2}{c}{Full model}  
\tabularnewline	

\multirow{-2}{*}{\begin{tabular}{c}  Training image \\\includegraphics[width=0.13\linewidth, height=0.06\textheight]{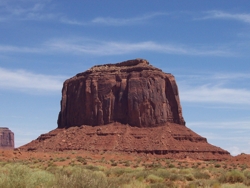} \end{tabular}} &
\includegraphics[width=0.13\linewidth, height=0.06\textheight]{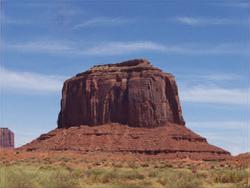} & 
\includegraphics[width=0.13\linewidth, height=0.06\textheight]{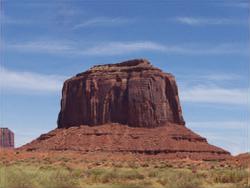} & 
\includegraphics[width=0.13\linewidth, height=0.06\textheight]{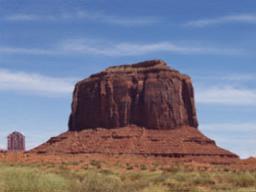} & 
\includegraphics[width=0.13\linewidth, height=0.06\textheight]{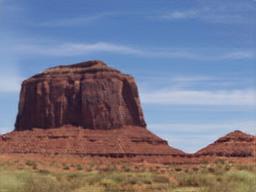} & 
\includegraphics[width=0.13\linewidth, height=0.06\textheight]{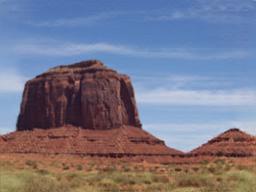} & 
\includegraphics[width=0.13\linewidth, height=0.06\textheight]{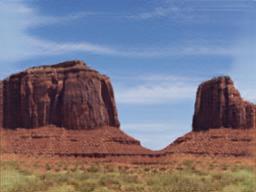}
\tabularnewline
&
\includegraphics[width=0.13\linewidth, height=0.06\textheight]{supplementary/figures/fa_dr_ablation/50_nodr/3} & 
\includegraphics[width=0.13\linewidth, height=0.06\textheight]{supplementary/figures/fa_dr_ablation/50_nodr/7} & 
\includegraphics[width=0.13\linewidth, height=0.06\textheight]{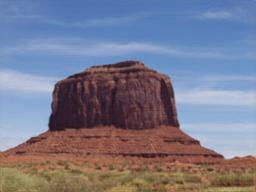} & 
\includegraphics[width=0.13\linewidth, height=0.06\textheight]{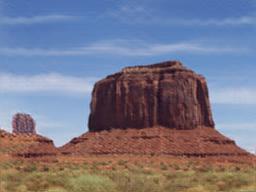} & 
\includegraphics[width=0.13\linewidth, height=0.06\textheight]{figures/bank_sin_im/50/15} & 
\includegraphics[width=0.13\linewidth, height=0.06\textheight]{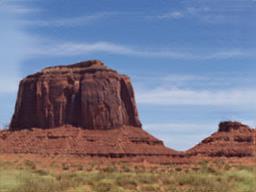}

\end{tabular}\hfill{}
\par\end{centering}
\vspace{-0.4em}
\caption{\label{fig:fa}  Qualitative ablation on the proposed diversity regularization (DR) and feature augmentation (FA). Without DR, the model does not mitigate memorization, producing only images that are perceptually indistinguishable from the original sample. Without FA, SIV-GAN achieves only modest diversity in content and layouts. Finally, using both DR and FA enables generating more interesting novel scene compositions like varying global scene layouts or duplicating and removing objects.}
\end{figure*} 

\begin{figure*}[t]
	\begin{centering}
		\setlength{\tabcolsep}{0.0em}
		\renewcommand{\arraystretch}{0}
		\par\end{centering}
	\begin{centering}
		\vspace{-0.9em}
		\begin{tabular}{c@{\hskip 0.03in}c@{\hskip 0.03in}c@{\hskip 0.03in}c@{\hskip 0.03in}c@{\hskip 0.03in}c}
		Training image  & $N_{D_{low-level}} = 1$ & $N_{D_{low-level}} = 2$ & $N_{D_{low-level}} = 3$ & $N_{D_{low-level}} = 4$ & $N_{D_{low-level}} = 5$ \tabularnewline			
	 		\includegraphics[width=0.16\linewidth, height=0.07\textheight]{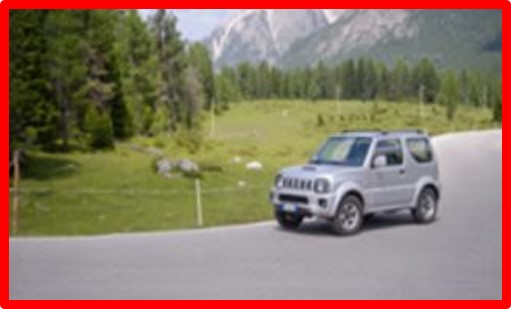} &
			\includegraphics[width=0.16\linewidth, height=0.07\textheight]{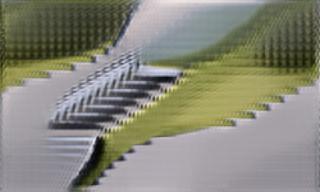} & 
			\includegraphics[width=0.16\linewidth, height=0.07\textheight]{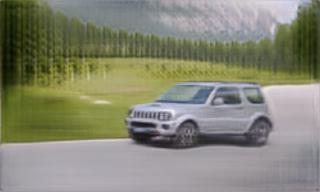} &
			\includegraphics[width=0.16\linewidth, height=0.07\textheight]{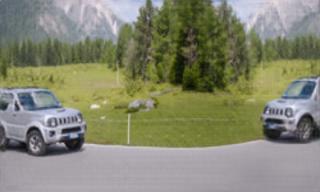} &
			\includegraphics[width=0.16\linewidth, height=0.07\textheight]{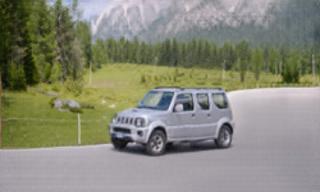} &
			\includegraphics[width=0.16\linewidth, height=0.07\textheight]{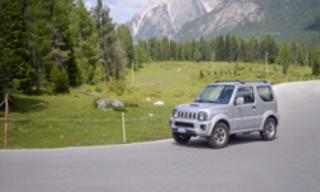}
		\end{tabular}\hfill{}
		\par\end{centering}
	\vspace{-0.9em}
	\caption{\label{fig:N_low_level} Effect of the number of discriminator blocks used before branching. Using too few blocks (1-2) leads to reduced image quality as $D_{low-level}$ is unable to extract the features necessary to build the content and layout representations. Increasing the number of blocks (3-4) results in improved quality while maintaining good diversity. Using too many blocks (e.g., 5 or more) leads to the memorization effect due to overfitting and the model tends to reproduce the training image with little diversity.}
	\vspace{-1.8em}
\end{figure*}

\begin{table}[t]
\setlength{\tabcolsep}{0.4em}
\renewcommand{\arraystretch}{1.00}
\centering
\caption{Comparison of diversity regularization techniques in the Single Image setting on DAVIS-YFCC100M.}

\begin{tabular}{c|c|c|c|c}
	\multirow{2}{*}{\normalsize{Regularization} } & \multirow{2}{*}{\normalsize{} SIFID~$\downarrow$} & \multirow{2}{*}{\small{} LPIPS~$\uparrow$ } & {\small{} Pixel ~$\uparrow$} & {\normalsize{} Dist.}  \tabularnewline
    & & & \small{} diversity & to train \tabularnewline
	\hline 	\hline 	
	{\normalsize None} &  \textbf{\small{0.05}}  &  \color{darkred} \small{0.04}  & \color{darkred} \small{0.33} & \small{\color{darkred} {0.06}}  \tabularnewline
	{\normalsize{} zCR } & \small{0.05}   &  \color{darkred} \small{0.06}  & \color{darkred} \small{0.37} &  \color{darkred} \small{0.09}  \tabularnewline
	{\normalsize{} DS  } & \small{0.06}  &  \color{darkred} \small{0.14} & \color{darkred} {\small{0.45}} &  \color{darkred} \small{0.14} \tabularnewline	
		{\normalsize{} DR (im. space)   } & \small{0.07}  &  \small{0.21} & {\small{0.52}} &  \small{0.25} \tabularnewline			
	{\normalsize{} DR   } & \small{0.08}  &  \textbf{\small{0.33}} & \textbf{{\small{0.66}}} & {{\small{0.37}}} \tabularnewline	
\end{tabular}
\label{table:ablation_diversity} %

\end{table}

Firstly, we ablate our discriminator architecture. We remove the layout branch, the content branch, or both (no branches). The latter corresponds to a standard GAN discriminator. 
The model without branches is trained together with our proposed diversity regularization (DR) and feature augmentation (FA), as well as differentiable augmentations (DA) as in \citep{Karras2020TrainingGA,Zhao2020DifferentiableAF}. However, as seen from Table \ref{table:main_ablation}, it memorizes the training images and reproduces them with poor diversity. Using only one of the branches shows good diversity, but the model fails to generate globally-coherent images, having a high $\frac{H\times W}{16}$ SIFID. The qualitative results for these ablation models in the Single Image setting are presented in Fig.~\ref{fig:qual_ablations_d}. We observe that the visual results correspond well to the conclusions from Table \ref{table:main_ablation}. For example, employing none of the branches only reproduces the training image. The model without the layout branch generates different objects in various combinations, but the model often fails to position the objects correctly or to generate globally-coherent layouts. In particular, there might be a horizon discontinuity, or air balloons may follow unrealistically structured positions in a grid. On the other hand, the model trained without the content branch generates images with more realistic layouts, but does not preserve the content distribution of the original training image, distorting the appearance of objects or perturbing their shapes.

Next, we observe the effect of the proposed DR and FA. Without DR, the model does not achieve multi-modality, scoring low in all the diversity metrics.
The absence of FA notably decreases diversity, resulting in the diversity scores dropping by 0.02-0.08 points. The qualitative results for these models are shown in Fig. \ref{fig:fa}. SIV-GAN without DR does not mitigate overfitting, suffering from mode collapse. The model with DR but without FA manages to generate diverse images. However, such a model produces only modest diversity in content and layouts, e.g., it only slightly translates a rock to new locations in the image.

Finally, removing the low-level loss $\mathcal{L}_{D_{low-level}}$ also results in decreased diversity. According to Eq.~\ref{eq:loss_d_1} and \ref{eq:loss_d_2}, this term shifts the attention of the loss function from the latest discriminator layers towards earlier layers with smaller receptive field, which impedes the memorization of the whole image. This way, $\mathcal{L}_{D_{low-level}}$ not only helps to learn low-level image statistics, but also to regularize the discriminator, which results in a higher diversity. 

\begin{table}
	\setlength{\tabcolsep}{0.4em}
	\renewcommand{\arraystretch}{1.00}
	\centering
	\caption{Effect of the diversity regularization (DR) strength in the Single Image setting on DAVIS-YFCC100M.}
	\begin{tabular}{c|c|c|c|c}
		\multirow{2}{*}{\normalsize{} $\lambda$} & \multirow{2}{*}{\normalsize{} SIFID~$\downarrow$} & \multirow{2}{*}{\normalsize{} LPIPS~$\uparrow$ } & {\normalsize{} \normalsize{} Pixel ~$\uparrow$}  & {\normalsize{} Dist.} \tabularnewline
		& & & diversity & to train \tabularnewline
		
		\hline 	\hline 	
		
		\normalsize{} {0.00} &  \small{} \textbf{{0.05}}  &  \small{} \color{darkred} {0.04} & \small{} \color{darkred} {0.33} & \small{} \color{darkred} {0.06} \tabularnewline
		
		\normalsize{} {{} 0.05   } & \small{} {0.07}  & \small{} {{0.19}}  & \small{} {{0.50}} &  \small{} {{0.23}} \tabularnewline
		
		\normalsize{} {{} 0.15   } & \small{} {0.08}  &  \small{} {{0.33}}  & \small{} {{0.66}} &  \small{} {0.37} \tabularnewline
	
		\normalsize{} { 0.50  } & \small{} \color{darkred} {0.13}  &  \small{} \textbf{{{0.39}}} & \small{} \textbf{{{0.69}}} & \small{} {{0.46}} \tabularnewline
		
	\end{tabular}
	\label{table:lambda_dr} %
\end{table}

\myparagraph{Comparison of DR to alternative techniques.}
In Table \ref{table:ablation_diversity}, we compare our proposed DR to the latent consistency regularization (zCR) \citep{Zhao2020ImprovedCR}, diversity-sensitive loss (DS) \citep{Yang2019DiversitySensitiveCG}, and using no regularization (none). We apply zCR only to the generator loss, leaving the discriminator objective as it is. 
As both zCR and DS operate in the image space, we also test our proposed DR in the image space instead of the $G$ feature space, as in \citep{choi2020stargan}. 
As seen from Table~\ref{table:ablation_diversity}, our DR noticeably improves over zCR and DS in all diversity metrics. Moreover, we find it beneficial to use DR in the feature space, which leads to more variation in the generated samples. Interestingly, Table~\ref{table:ablation_diversity} illustrates a quality-diversity trade-off in the Single Image setting, where improvements in diversity lead to deterioration in SIFID.

\myparagraph{Ablation on the DR strength.} In all our experiments, we used DR with $\lambda{=}0.15$. In Table \ref{table:lambda_dr}, we show the effect of changing $\lambda$ for DR in the Single Image setting, and thus changing the strength of the diversity regularization. 
Table \ref{table:lambda_dr} shows that setting the weight too high ($\lambda{=}0.50$) leads to good diversity but harms image quality, whereas small values ($\lambda{=}[0.00, 0.05]$) are beneficial for quality but deteriorates diversity. We observed that using $\lambda {=} 0.15$ leads to a good trade-off, resulting in a high diversity among generated samples while not corrupting the quality of textures and the global layout coherency.

\begin{table}
	\setlength{\tabcolsep}{0.4em}
	\renewcommand{\arraystretch}{1.00}
	\centering

	\caption{Ablation on the number of blocks $N_{D_{low-level}}$ used before the content-layout branching on the DAVIS-YFCC100M dataset. }

	\begin{tabular}{c|c|c||c|c}
		 \multirow{2}{*}{\normalsize $N_{D_{low-level}}$} & \multicolumn{2}{c||}{\small Single Image}  & \multicolumn{2}{c}{\small Single Video}  
		\tabularnewline
		  & {{} \small SIFID~$\downarrow$}  & {{} \small LPIPS~$\uparrow$ } &  {{} \small SIFID~$\downarrow$} & {{} \small LPIPS~$\uparrow$ } \tabularnewline

		\hline 	\hline 	
		
		 \normalsize 1 & \small \color{darkred} {{0.59}}  &  \small \textbf{{0.42}} &  \small \color{darkred} {2.75} &  \small \textbf{{0.46}}  \tabularnewline
		
		 \normalsize 2 & \small \color{darkred} {{0.13}}  &  \small {0.40} &  \small \color{darkred} {1.12} &  \small {0.45}  \tabularnewline
		 
		 \normalsize 3 & \small {{0.08}}  &  \small {0.33} & \small  {0.55} & \small  {0.43}  \tabularnewline

		 \normalsize 4 & \small {{0.06}}  &  \small {0.24} &  \small {0.36} &  \small {0.38}  \tabularnewline

		\normalsize 5 & \small \textbf{{0.03}}  &  \small \color{darkred} {0.15} &  \small {0.40} &  \small {0.37}  \tabularnewline

		\normalsize 6 & \small \textbf{{0.03}}  &  \small \color{darkred} {0.13} &  \small \textbf{{0.35}} &  \small {0.36}  \tabularnewline

		\end{tabular}

\label{table:N_low_level} %

\end{table}

\begin{figure*}[t]
	\begin{centering}
		\setlength{\tabcolsep}{0.0em}
		\renewcommand{\arraystretch}{0}
		\par\end{centering}
	\begin{centering}
	\begin{tabular}{c@{\hskip 0.05in}c@{\hskip 0.05in}c@{\hskip 0.05in}}
			
			 & \multicolumn{2}{c}{Generated samples from a single training image at a resolution of 512x896}
			\tabularnewline 
			
			\multirow{-4}{*}{\begin{tabular}{c}  Training image (512x896) \\ 	\includegraphics[width=0.25\linewidth]{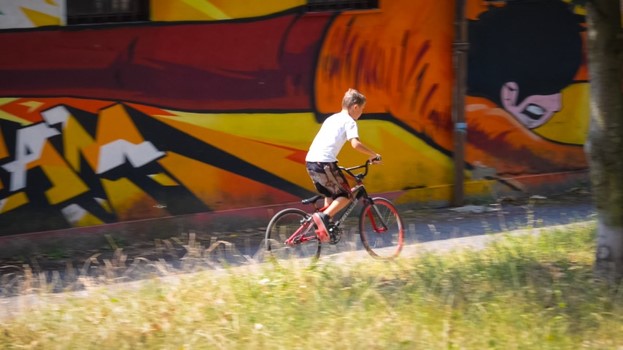}\end{tabular}  } & 	

			\includegraphics[width=0.35\linewidth]{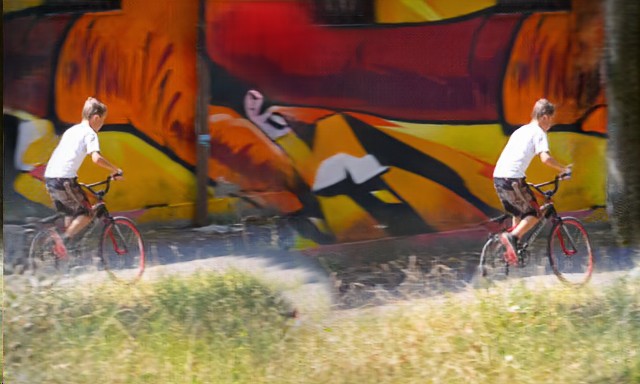} &
			\includegraphics[width=0.35\linewidth]{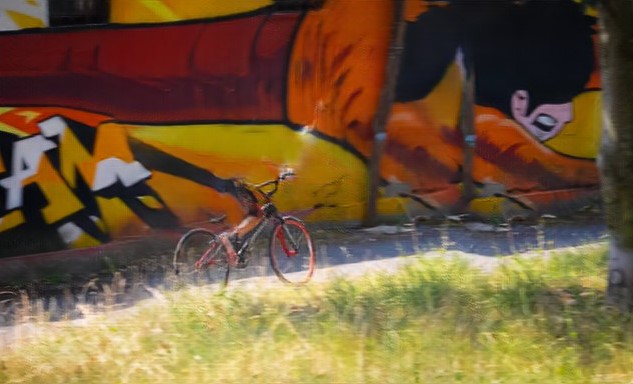} 
			\tabularnewline 
			& 
			\includegraphics[width=0.35\linewidth]{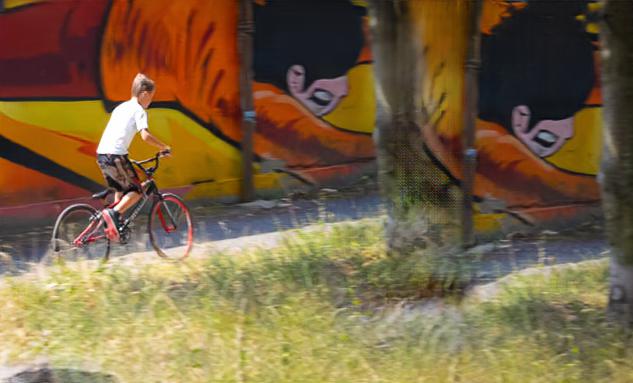} &
			\includegraphics[width=0.35\linewidth]{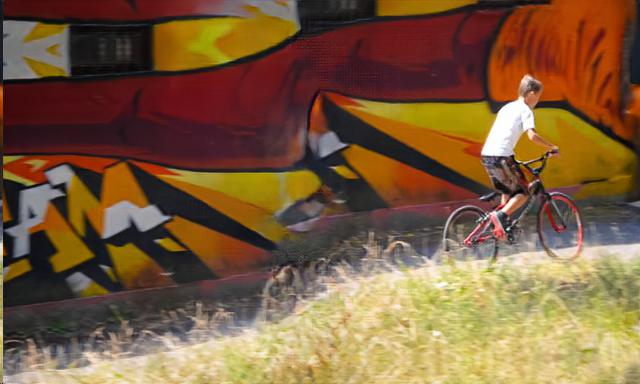} 
			
		\end{tabular}
		\par\end{centering}
	\caption{SIV-GAN results at a high image resolution of 512x896 in the Single Image setting. Our model shows good scalability to high image resolutions, maintaining the quality and diversity of scene compositions. For example, given only one image with a bike rider, SIV-GAN can produce images with two bikers, without the biker, as well as change the bike lane path.}
	\label{fig:high_res}
\end{figure*}

\myparagraph{Ablation on the number of low-level ResNet blocks.} 
The SIV-GAN discriminator uses 3 ResNet blocks for $D_{low-level}$ before the branching.
In Table \ref{table:N_low_level} and Fig.~\ref{fig:N_low_level}, we analyse the effect of applying branching at an earlier or a later discriminator stage, keeping the overall depth of the network equal to 7 ResNet blocks. The results indicate that the branching should be applied neither too early nor too late. Using too few ResNet blocks (1-2) before the branching leads to a reduced capacity of the low-level feature extractor $D_{low-level}$, so this network becomes unable to learn meaningful content and layout features. Such a model learns the color distribution of an image, but cannot produce a globally-coherent scene and generate textures of good quality as shown in Fig.~\ref{fig:N_low_level}. This effect is indicated by a very high SIFID in Table \ref{table:N_low_level}. On the other hand, using too many blocks before the branching, such as 5 or more, increased the capacity of $D_{low-level}$, so it becomes easier to memorize the whole image as shown in Fig.~\ref{fig:N_low_level} and by the low diversity metrics in Table~\ref{table:N_low_level}. We found that using $N_{low-level}{=}3$ leads to an optimal quality-diversity trade-off in both the Single Image and Single Video settings.




\begin{table}[t]
	\setlength{\tabcolsep}{0.4em}
	\renewcommand{\arraystretch}{1.30}
	\centering
	\caption{Comparison of synthesis quality and diversity at different image resolutions on DAVIS-YFCC100M in the Single Image setting.}

	\begin{tabular}{c|c|c}

		  \normalsize Image resolution & {\small SIFID~$\downarrow$}  & {\small LPIPS~$\uparrow$ }  \tabularnewline

		\hline 	\hline

		 \normalsize 192x320 & \small {{0.08}}  &  \small {{0.33}} \tabularnewline
		 \normalsize 512x896 & \small {{0.08}}  &  \small {{0.29}}

		\end{tabular}
\label{tab:high_res} %

\end{table}

\myparagraph{Effect of using a higher image resolution.}
In Fig. \ref{fig:high_res} and Table \ref{tab:high_res} we demonstrate the ability of our model to generate images at a higher image resolution of 512x896 on the DAVIS-YFCC100M dataset. For this, we add one ResNet block to the generator and discriminator, and change the input noise shape from 3x5 to 4x7. After this change, the model produces images at a much higher image resolution of 896x512. We show the visual results for high resolution image synthesis of SIV-GAN in Fig. \ref{fig:high_res}. We do not observe any issues caused by the change of image resolution. As shown in Table \ref{tab:high_res}, the performance of the model is similar at different scales: SIFID of 0.08 and LPIPS of 0.29 at resolution 512x896 is aligned well with quality and diversity at resolution 192x320 (0.08 and 0.33 in Table \ref{table:comp_single_image}). 

%
%
\section{Conclusion}
\label{sec:conclusion}

In this work, we explored uncondtional GAN training in extremely limited data regimes. We introduced a new task of learning generative models from a single video, which presents a new challenge for few-shot image synthesis models due to a high similarity between training images. We proposed SIV-GAN, a new model that successfully learns from a single image or a single video, outperforming prior work.
In such extremely low-data regimes, our model prevents memorization and generates diverse images that are significantly different from the training set. Inherently, the synthesis of our model is constrained by the appearance of objects present in the original sample. Nevertheless, SIV-GAN can synthesize novel scene compositions by blending objects in different combinations, changing their shape or position, while preserving the original context and plausibility of the scene. Notably, such compositionality is enabled by the model's ability to distinguish objects and backgrounds learnt just from a single image or video. In \citep{sushko2023wacv}, we have already demonstrated that SIV-GAN is a useful tool for data augmentation in domains where data collection remains challenging.

\section*{Acknowledgement}

The work has been supported by the ERC Consolidator Grant FORHUE  (101044724).

\bibliographystyle{model2-names}
\bibliography{references}

\end{document}